\documentclass[letterpaper, 10 pt, conference]{ieeeconf}  % Comment this line out if you need a4paper

\IEEEoverridecommandlockouts                              % This command is only needed if 
\usepackage[utf8]{inputenc} % allow utf-8 input
\usepackage[T1]{fontenc}    % use 8-bit T1 fonts

\usepackage{url}            % simple URL typesetting

\usepackage{breakurl}
\usepackage{hyperref}       % hyperlinks
\usepackage{nicefrac}       % compact symbols for 1/2, etc.
\usepackage{microtype}      % microtypography
\usepackage{xcolor}         % colors
\usepackage[dvipsnames]{xcolor}
\usepackage{graphicx, wrapfig, blindtext}
\usepackage{multirow}
\usepackage{multicol}
\usepackage{makecell}
\usepackage{subcaption}
\usepackage{bibunits}

\usepackage{booktabs}   %% For formal tables:
\usepackage{latexsym}
\usepackage{xspace}
\usepackage{amsmath}

\usepackage{amssymb}  % assumes amsmath package installed

\usepackage[rightcaption]{sidecap}
\usepackage{changepage}
\usepackage{cancel}
\usepackage{tabularx,colortbl, hhline}
\usepackage{listings}
\usepackage{verbdef}
\usepackage{moreverb}
\usepackage{fancyvrb}
\usepackage{framed}
\usepackage{textcomp}
\usepackage{noindentafter}
\usepackage{atbegshi,ifthen,tikz}
\usepackage[flushleft]{threeparttable}
\usepackage{marginnote}
\usepackage{color}
\usepackage{balance}
\usepackage{flushend}
\usepackage{soul}

\usepackage{graphics} % for pdf, bitmapped graphics files
\usepackage{epsfig} % for postscript graphics files
\usepackage{times} % assumes new font selection scheme installed

\usepackage{lipsum}
\usepackage{diagbox}
\usepackage{courier}
\usepackage[english]{babel}
\usepackage{seqsplit}
\usepackage{hyperref}
\hypersetup{
    colorlinks=true,
    linkcolor=blue,
    citecolor=blue,
    filecolor=blue,      
    urlcolor=blue,
}
\usepackage{cite}
\usepackage[noend]{algpseudocode}
\usepackage{algorithm}
\usepackage{amsmath,amsfonts,bm}
\usepackage{setspace}

\usepackage{pifont}

\def\eqref#1{equation~\ref{#1}}
\def\1{\bm{1}}

\DeclareMathAlphabet{\mathsfit}{\encodingdefault}{\sfdefault}{m}{sl}
\SetMathAlphabet{\mathsfit}{bold}{\encodingdefault}{\sfdefault}{bx}{n}

\definecolor{green}{rgb}{0.0, 0.5, 0.0}

\makeatletter

        \newcount\bt@rangea
\newcount\bt@rangeb

\newcommand\btIfInRange[2]{%
	\global\let\bt@inrange\@secondoftwo%
	\edef\bt@rangelist{#2}%
	\foreach \range in \bt@rangelist {%
		\afterassignment\bt@getrangeb%
		\bt@rangea=0\range\relax%
		\pgfmathtruncatemacro\result{ ( #1 >= \bt@rangea) && (#1 <= \bt@rangeb) }%
		\ifnum\result=1\relax%
		\breakforeach%
		\global\let\bt@inrange\@firstoftwo%
		\fi%
	}%
	\bt@inrange%
}

\newcommand\bt@getrangeb{%
	\@ifnextchar\relax%
	{\bt@rangeb=\bt@rangea}%
	{\@getrangeb}%
}

\def\@getrangeb-#1\relax{%
	\ifx\relax#1\relax%
	\bt@rangeb=100000%   \maxdimen is too large for pgfmath
	\else%
	\bt@rangeb=#1\relax%
	\fi%
}

\newcommand{\btLstHL}[1]{%
	\btIfInRange{\value{lstnumber}}{#1}%
	{\color{light-gray}}%
	{\def\lst@linebgrd}%
}%

\lst@Key{numbers}{none}{%
    \def\lst@PlaceNumber{\lst@linebgrd}%
    \lstKV@SwitchCases{#1}%
    {none:\\%
     left:\def\lst@PlaceNumber{\llap{\normalfont
        \lst@numberstyle{\thelstnumber}\kern\lst@numbersep}\lst@linebgrd}\\%
     right:\def\lst@PlaceNumber{\rlap{\normalfont
                \kern\linewidth \kern\lst@numbersep
                \lst@numberstyle{\thelstnumber}}\lst@linebgrd}%
    }{\PackageError{Listings}{Numbers #1 unknown}\@ehc}
}

\lst@Key{lbc}{}{%
	\def\lst@linebgrdcolor{#1}%
}
\lst@Key{linebackgroundsep}{0pt}{%
	\def\lst@linebgrdsep{#1}%
}
\lst@Key{linebackgroundwidth}{\linewidth}{%
	\def\lst@linebgrdwidth{#1}%
}
\lst@Key{linebackgroundheight}{\ht\strutbox}{%
	\def\lst@linebgrdheight{#1}%
}
\lst@Key{linebackgrounddepth}{\dp\strutbox}{%
	\def\lst@linebgrddepth{#1}%
}
\lst@Key{linebackgroundcmd}{\color@block}{%
	\def\lst@linebgrdcmd{#1}%
}

\newcommand{\lst@linebgrd}{%
	\ifx\lst@linebgrdcolor\empty\else
	\rlap{%
		\lst@basicstyle
		\color{-.}% By default use the opposite (`-`) of the current color (`.`) as background
		\lst@linebgrdcolor{%
			\kern-\dimexpr\lst@linebgrdsep\relax%
			\lst@linebgrdcmd{\lst@linebgrdwidth}{\lst@linebgrdheight}{\lst@linebgrddepth}%
		}%
	}%
	\fi
}

\makeatother

\definecolor{light-gray}{gray}{0.85}

\makeatletter
\let\origthelstnumber\thelstnumber

\newcommand*\Suppressnumber{%
	\lst@AddToHook{OnNewLine}{%
		\let\thelstnumber\relax%
		\advance\c@lstnumber-\@ne\relax%
	}%
}

\newcommand*\Reactivatenumber[1]{%
	\setcounter{lstnumber}{\numexpr#1-1\relax}
	\lst@AddToHook{OnNewLine}{%
		\let\thelstnumber\origthelstnumber%
		\refstepcounter{lstnumber}
	}%
}

\makeatother

{\endMakeFramed}

\newcommand\mycomment[1]{}

\newcommand{\toolname}{{\sc EffiTune}\xspace}

\title{%\LARGE 
\bf
\toolname:  
Diagnosing and Mitigating Training Inefficiency \\ for Parameter Tuner in Robot Navigation System
}

\author{Shiwei Feng, Xuan Chen, Zikang Xiong, Zhiyuan Cheng, \\ Yifei Gao, Siyuan Cheng, Sayali Kate and Xiangyu Zhang 
\thanks{All authors are affiliated with Department of Computer Science, Purdue University, West Lafayette, IN 47906, USA. Email: \texttt{\big\{feng292, chen4124, xiong84, cheng443, gao749, cheng535, skate, xyzhang\big\}@purdue.edu}
}% <-this % stops a space
}

\begin{document}
\maketitle
\thispagestyle{empty}
\pagestyle{empty}

%%%%%%%%%%%%%%%%%%%%%%%%%%%%%%%%%%%%%%%%%%%%%%%%%%%%%%%%%%%%%%%%%%%%%%%%%%%%%%%%

\begin{abstract}
Robot navigation systems are critical for various real-world applications such as delivery services, hospital logistics, and warehouse management. Although classical navigation methods provide interpretability, they rely heavily on expert manual tuning, limiting their adaptability. Conversely, purely learning-based methods offer adaptability but often lead to instability and erratic robot behaviors. Recently introduced parameter tuners aim to balance these approaches by integrating data-driven adaptability into classical navigation frameworks. However, the parameter tuning process currently suffers from training inefficiencies and redundant sampling, with critical regions in environment often underrepresented in training data. \looseness=-1

In this paper, we propose \toolname, a novel framework designed to diagnose and mitigate training inefficiency for parameter tuners in robot navigation systems. \toolname first performs robot-behavior-guided diagnostics to pinpoint critical bottlenecks and underrepresented regions. It then employs a targeted up-sampling strategy to enrich the training dataset with critical samples, significantly reducing redundancy and enhancing training efficiency. Our comprehensive evaluation demonstrates that \toolname achieves more than a 13.5\% improvement in navigation performance, enhanced robustness in out-of-distribution scenarios, and a 4× improvement in training efficiency within the same computational budget. 

\end{abstract}

\section{Introduction} 
\label{sec:introduction}

Robot navigation is essential for applications like delivery services, hospital logistics, warehouse management, and library automation. Navigating cluttered environments remains a key challenge for achieving agile, efficient, and safe movement. Classical navigation systems~\cite{abdulsaheb2022robot, asano1985visibility, Canny1985AVM, Sampling2014} address this by using specific cost functions and rules, which are difficult to generalize and often require expert manual tuning~\cite{Campbell2020, raheem2022robot, zheng2021rostuning, feng2024rocas,kate2025roscallbax}.
In contrast, learning-based approaches~\cite{pfeiffer2017perception, gao2017intention, chiang2019learning, xiao2021toward, liu2021lifelong, xiao2021agile, xiao2020appld} offer adaptability but can suffer from instability and lack interpretability~\cite{feng2023decree, chen2024temporalrl,cheng2024fusion,cheng2024badpart}, leading to erratic behaviors that compromise safety and efficiency.

\begin{figure}[t]
    \centering
    \begin{subfigure}[t]{.48\linewidth}
        \centering
        \includegraphics[width=1\textwidth]{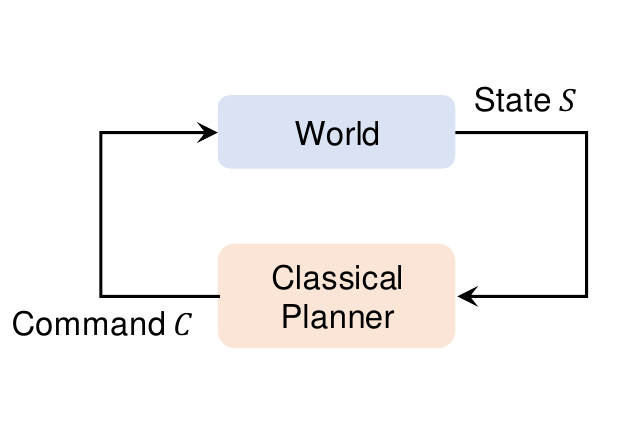}
        \caption{Classical}
        \label{fig:classical_planner}
    \end{subfigure}
    % \hspace{1pt}
    % \begin{subfigure}[t]{.45\linewidth}
    %     \centering
    %     \includegraphics[width=1.1\textwidth]{figs/intro-2.pdf}
    %     \caption{\small RL End-to-End }
    %     \label{fig:rl_planner}
    % \end{subfigure}
    % \hspace{1pt}
    \begin{subfigure}[t]{.49\linewidth}
        \centering
        \includegraphics[width=1\textwidth]{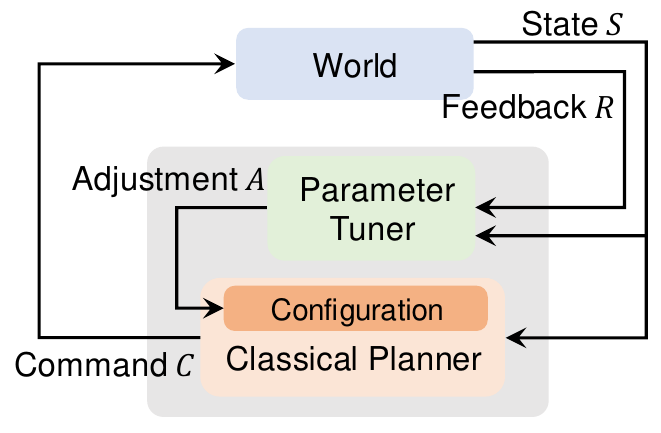}
        \caption{Hybrid }
        \label{fig:hybrid_planner}
    \end{subfigure}
    \caption{Different Types of Robot Navigation Systems. Parameter tuners are typically data-driven and requires training.}
    \label{fig:types_planner}
    \vspace{-15pt}
\end{figure}

To overcome these limitations, recent studies introduced parameter tuner~\cite{xu2021applr, xiao2022motion,jiang2022efficient,jiang2023learning} that integrate data-driven tuner with classical planning frameworks (see Fig.~\ref{fig:types_planner}). These data-driven parameter tuners adjust the parameters (across perception, planning and control components) of classical navigation systems in different scenarios, combining adaptability of data-driven methods with the explainable decision-making of classical methods for balanced robot navigation.

\indent However, existing parameter tuners encounter significant challenges. The interaction between classical navigation systems and the environment is computationally expensive, often leading to sampling inefficiencies. Furthermore, the training process of these tuners tends to include redundant sampling.
In addition, our analysis reveals that only a few key regions in the environment are critical for achieving optimal performance. Unfortunately, these pivotal areas are underrepresented in the training data due to both the high cost of environmental interaction and the prevalence of redundant sampling.
To address these challenges, our approach emphasizes the most informative data during training, significantly enhancing the effectiveness and efficiency of the parameter tuner. Unlike static roadmap analysis, which offers a fixed representation, our method captures the robot's dynamic behavior, providing a more accurate and informative analysis. \looseness=-1

\indent In this paper, we propose \toolname, a diagnosis and mitigation framework that addresses inefficiency in parameter tuner training. \toolname begins with a robot-behavior-guided diagnostic phase that identifies critical bottlenecks. It then employs a novel up-sampling strategy to selectively enrich the training data with samples from these key regions, providing the parameter tuner with a more balanced and informative dataset while reducing redundancy. This targeted approach accelerates the training process of tuners, enhances adaptability, and ultimately improves overall navigation performance. \looseness=-1

In summary, our contributions are: \looseness=-1
\begin{itemize}
     \item We identify training inefficiency issue for parameter tuner training in robot navigation. \looseness=-1
     \item We introduce \toolname, a framework that features a diagnostic phase to detect these critical regions and a mitigation phase that uses targeted data up-sampling to accelerate parameter tuner training. \looseness=-1
     \item \toolname achieves a 13.5\%+ improvement in navigation performance, greater robustness in out-of-distribution environments, and a 4× increase in training efficiency within the same timing budget. 
\end{itemize}

\section{Background, Notations and Related Work} 
\label{sec:related}

\subsection{Motion Planning} \label{sec:related:motion-planning}

\begin{figure}[t]
    \centering
    \begin{subfigure}[t]{.15\textwidth}
        \centering
        \includegraphics[width=1\textwidth]{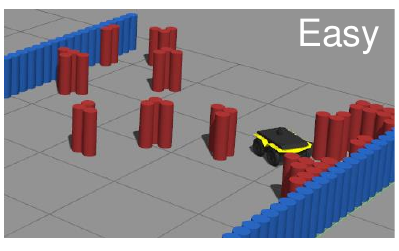}
        \caption{0 $\sim$ 2.5 s/m}
        \label{fig:gazebo_3d_easy}
    \end{subfigure}
    % \hfill
    % \nextfloat
    \begin{subfigure}[t]{.15\textwidth}
        \centering
        \includegraphics[width=1\textwidth]{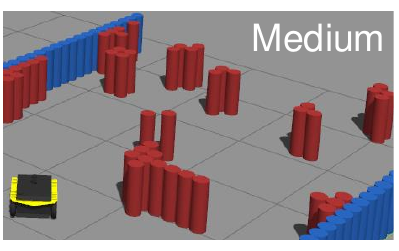}
        \caption{2.5 $\sim$ 4.0 s/m}
        \label{fig:gazebo_3d_mid}
    \end{subfigure}
    % \hfill
    % \nextfloat
    \begin{subfigure}[t]{.15\textwidth}
        \centering
        \includegraphics[width=1\textwidth]{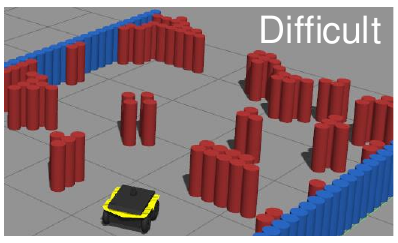}
        \caption{4.0 $\sim$ 8.0 s/m}
        \label{fig:gazebo_3d_difficult}
    \end{subfigure}
    \hfill
    \caption{Maps of various difficulty levels. We use the difficulty metric of ``normalized traversal time'' (traversal time averaged over the ten trials and normalized by path length) from ~\cite{perille2020benchmarking} to partition map datasets to 3 levels.}
    \label{fig:gazebo_3d}
    \vspace{-8pt}
\end{figure}

{\it Motion planning} is used to determine a trajectory for a robot 
to move from a start position to a goal position while avoiding obstacles (as visualized in Fig.~\ref{fig:gazebo_3d}). 
There are 2 major categories of planning algorithms: local trajectory planners\cite{fox1997dwa,teb} and global path planners~\cite{dijkstra, astar, lavalle1998rrt, lavalle2001rrt}. \looseness=-1

In this paper, we consider motion planning in navigation tasks and assume that the robot employs a classical \textit{local} motion planner $f$.
The local planner $f$ can be tuned via a set of configurations (i.e., configurable parameters) $\theta \in \Theta$, where $\Theta$ denotes all possible values of planner configuration.
While navigating in a physical world $\mathcal{W}$ with obstacles, $f$ tries to move the robot to a local goal $g_t \in \mathbb{R}^2$ (computed by the global planner).
At each time step $t$, $f$ receives sensor observations $o_t$ (e.g. lidar scans) to do collision avoidance, and then attempts to reach the local goal $g_t$ in a fast and collision-free manner.
The local planner $f$ is responsible for computing the motion commands $u_t=f(o_t, g_t \vert \theta)$ (e.g., angular or linear velocity) at each time step to reach $g_t$.

\subsection{Parameter Tuner} \label{sec:related:meta-planner}

\setlength{\intextsep}{0pt}

\textit{Parameter Tuner} $\tau$ operates on top of the original planner $f$, taking environmental states $s$ as input and adjusting navigation configuration $\theta$. Fig.~\ref{fig:types_planner}c shows that a hybrid planner with a parameter tuner combines the explainability of classical planners with the adaptability of data-driven (e.g., end-to-end) planners, enhancing performance. It has been applied across domains such as robot navigation~\cite{fridovich2018planning, xiao2022motion, siva2019robot, pokle2019deep}, self-driving~\cite{jiang2022efficient}, autonomous system~\cite{feng2025intentest,zheng2025llm}, drone control~\cite{loquercio2022autotune}, and robotic manipulation~\cite{Moll2021HyperPlanAF, jia2023dynamic}.\looseness=-1

There are two types of parameter tuners: static and dynamic. Static parameter tuners tune the configuration $\theta$ before deployment, while dynamic parameter tuners adjust it in real-time.

\noindent {\bf Static.} Bayesian optimization is often used to auto-tune parameters for black-box systems~\cite{snoek2012practical, shahriari2015taking, frazier2018tutorial}. By iteratively updating its model of the objective function, it efficiently explores the parameter space, tuning planning parameters for better performance.

\noindent {\bf Dynamic.} Dynamic parameter tuners adjust parameters on-the-fly for different scenarios~\cite{xiao2021learning, Binch2020, xiao2020appld, xu2021applr}. For example, \cite{xiao2020appld} learns a library of parameter sets and switches based on context, while \cite{xu2021applr} uses deep neural network to optimize parameters under various situations.

\subsection{Roadmap Analysis}
Existing roadmap analysis~\cite{kavraki1998roadmapanalysis, hsu2006foundationroadmap} often relies on static geometric properties to address challenges like narrow passages, using concepts such as visibility and lookout.
While these studies provide valuable insights, they do not fully capture the dynamic execution status resulting from a robot's interaction with environment.
Our approach aims to enhance performance by leveraging robot behavior during dynamic execution. By analyzing the robot's behaviors as it operates, we identify high-resistance areas that may not be apparent through static analysis alone. This behavior-guided method allows us to adaptively improve planning and navigation, leading to better performance than static geometric analysis.

\section{Training Inefficiency of Existing Methods} 
\label{sec:existing}

APPLR~\cite{xu2021applr}, the state-of-the-art dynamic parameter tuner, exhibits a training inefficiency that can lead to limited performance improvements. In this section, we discuss the training inefficiency, key insights, and related technical challenges. \looseness=-1

\begin{figure*}[t]
    \centering
    \begin{subfigure}[t]{.245\textwidth}
        \centering
        \includegraphics[width=1\textwidth]{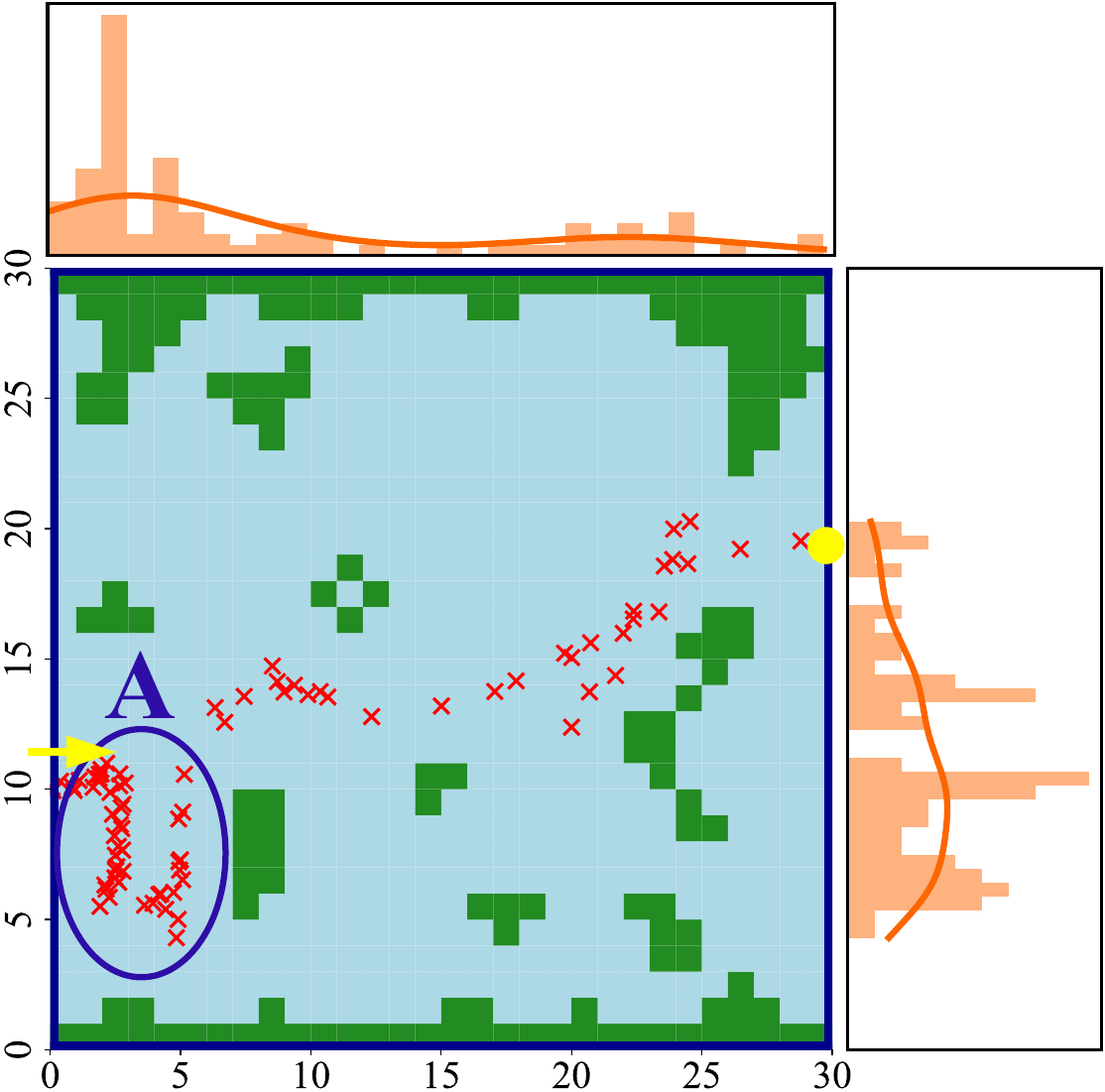}
        \caption{\small Iteration-1}
        \label{fig:moti_a}
    \end{subfigure}
    \hfill
    % \nextfloat
    \begin{subfigure}[t]{.245\textwidth}
        \centering
        \includegraphics[width=1\textwidth]{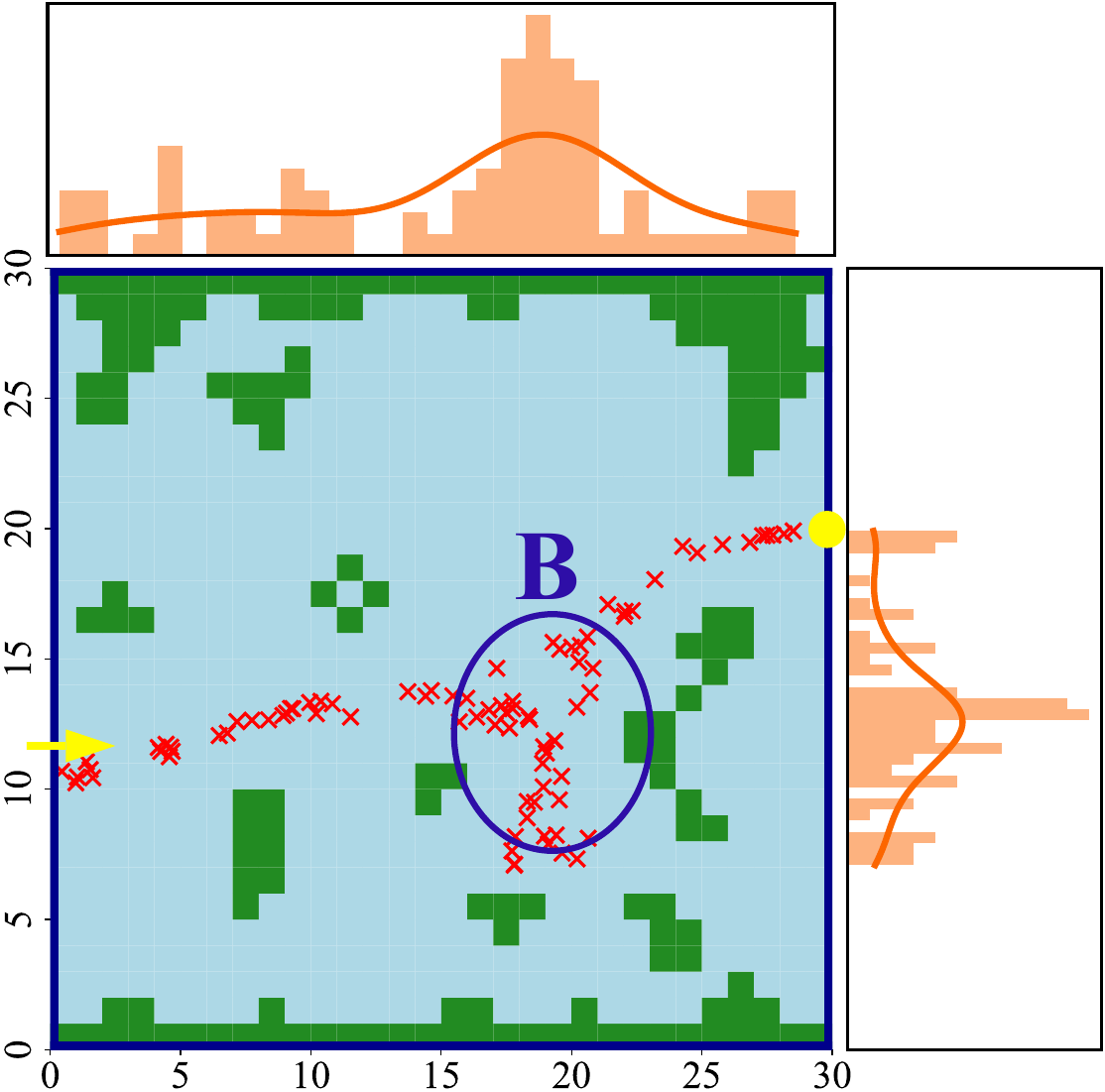}
        \caption{\small Iteration-10}
        \label{fig:moti_b}
    \end{subfigure}
    \hfill
    % \nextfloat
    \begin{subfigure}[t]{.245\textwidth}
        \centering
        \includegraphics[width=1\textwidth]{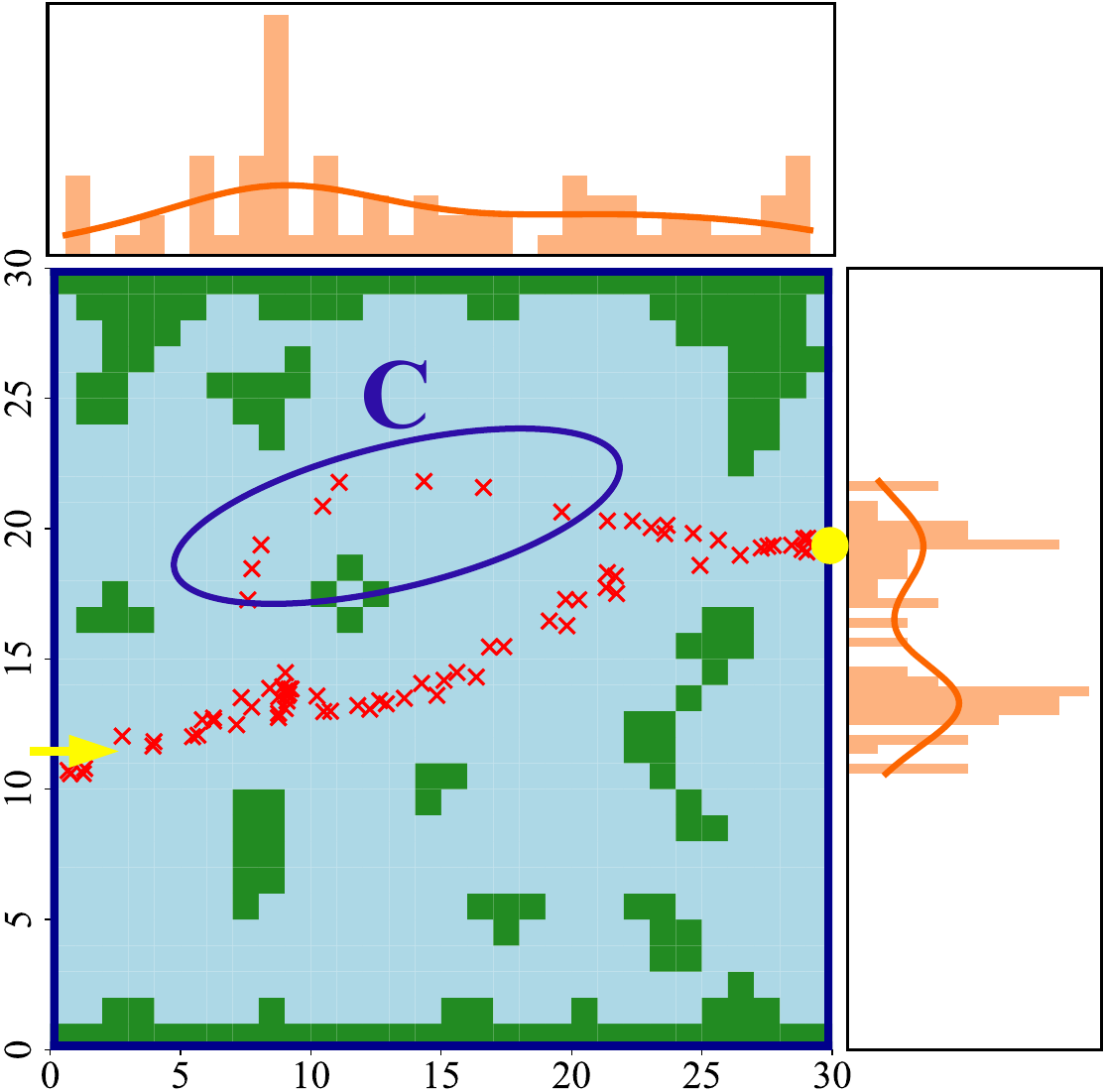}
        \caption{\small Iteration-30}
        \label{fig:moti_c}
    \end{subfigure}
    \hfill
    % \nextfloat
    \begin{subfigure}[t]{.245\textwidth}
        \centering
        \includegraphics[width=1\textwidth]{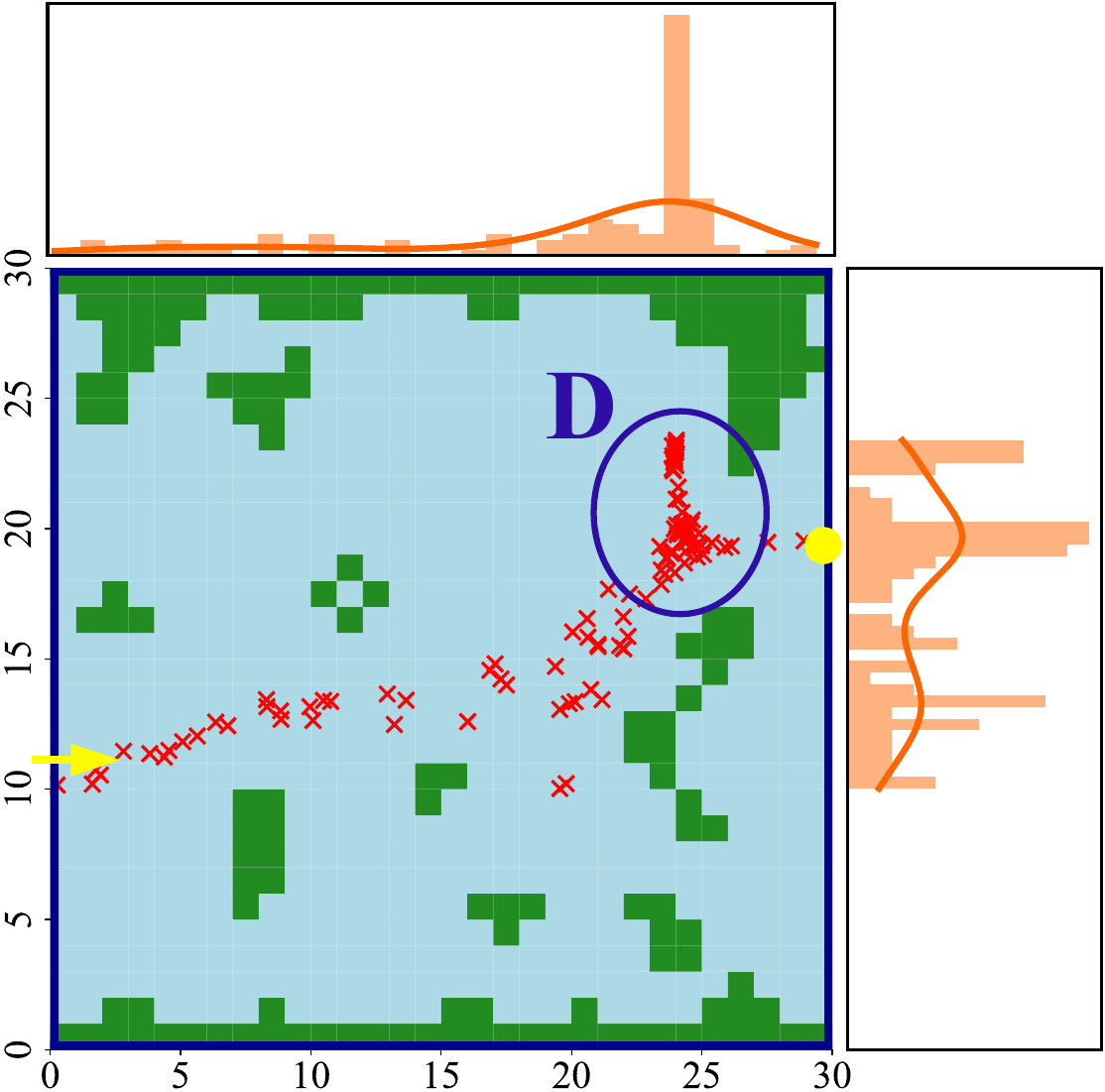}
        \caption{\small Iteration-90}
        \label{fig:moti_d}
    \end{subfigure}
    \hfill
    % \nextfloat
    \caption{Training inefficiency in existing parameter tuner. For convenient visualization, we show 2D projections of the difficult map from Fig.~\ref{fig:gazebo_3d} and omit the robot body. 
    The \textbf{\color{Goldenrod}{yellow arrow}} denotes the start position and the \textbf{\color{Goldenrod}{yellow circle}} denotes the goal. 
    \textbf{\color{Red}Red cross} markers are the robot trajectories at each timestamp (each subfigure contains 3 trajectories to show the distribution). On top and right of the maps are the frequency distribution of robot positions $x$ and $y$.
    }
    \label{fig:moti}
\end{figure*}

\noindent {\bf Parameter Tuner Training.} 
Fig.~\ref{fig:moti} shows 2D projections of the \textit{difficult} map from Fig.~\ref{fig:gazebo_3d}, where the robot navigates obstacles from the start (left) to the goal (right) without collisions and within a time limit. The parameter tuner adjusts planner's configurations, i.e. adjustment $a_t = \tau(s_t)$, at a fixed frequency (e.g., 1~Hz) in response to environmental changes, and we record the robot's positions for reward calculation and tuner training. \looseness=-1

Fig.~\ref{fig:moti_a} shows the early-stage challenges with a randomly initialized tuner, where the robot often gets stuck (Area A). Over time (Fig.~\ref{fig:moti_b}), the tuner improves, avoiding early obstacles but still struggling with tight spaces. As training progresses (Fig.~\ref{fig:moti_c}), the robot reduces getting stuck but takes inefficient detours (Area C), leading to timeout. Further training (Fig.~\ref{fig:moti_d}) improves navigation, but the robot still faces difficulties passing through narrow gates (Area D).

\noindent {\bf Training Inefficiency.} From training iterations in Fig.~\ref{fig:moti}, training inefficiency can be summarized by 3 observations:\looseness=-1

\noindent \underline{\textit{Observation I}}: 
\textit{Challenging scenarios require more sampling}. In situations like dead corners (Fig.~\ref{fig:moti_b}), detours (Fig.~\ref{fig:moti_c}), and narrow gates (Figures~\ref{fig:moti_a}, \ref{fig:moti_d}), the classical planner's limitations and insufficient sampling of the parameter tuner often cause delays, timeouts, or collisions. These critical \textit{hard-to-pass} areas require significantly more sampling to improve navigation efficiency and safety.

\noindent \underline{\textit{Observation II}}: \textit{Challenging scenarios prevent training on subsequent scenarios}. 
\textit{Hard-to-pass} areas make the following scenarios (i.e., \textit{hard-to-reach}) more difficult to access. This is reflected in the robot's position distribution, showing higher frequency near \textit{hard-to-pass} areas (e.g., Area A in Fig.~\ref{fig:moti_a} and Area B in Fig.~\ref{fig:moti_b}) and lower frequency in subsequent regions. \looseness=-1

\noindent \underline{\textit{Observation III}}: \textit{It is hard and expensive to train on challenging scenarios}. 
Certain challenging scenarios remain under-fitted, risking mission success due to two key factors. For instance, Area D initially remains \textit{hard-to-reach} (Fig.~\ref{fig:moti_a}-\ref{fig:moti_c}) and is not accessible to the robot. As training progresses (Fig.~\ref{fig:moti_d}), Area D becomes accessible, as the robot can reach it smoothly but takes around 20 steps (about 20 seconds) per trajectory to arrive in Area D. This lengthy process is unnecessary but expensive and makes the tuner difficult to sufficiently train on Area D, leading to inefficient training within the given time budget.

\noindent {\bf Our Insight.} 
Observations \textit{I} and \textit{II} highlight the importance of accurately identifying challenging regions and adequately training within them. However, \textit{Observation III} reveals that extensive training in these areas is computationally costly, necessitating efficiency improvements.

To address these issues, we propose a two-stage approach: (1) leveraging a \textit{robot-behavior-guided} method to pinpoint critical and challenging regions effectively, and (2) employing targeted up-sampling strategies to reduce the overhead associated with sampling initiation. In Section~\ref{sec:design}, we detail our methodology for identifying these critical regions, directly addressing the inefficiency issue in parameter tuner training.

\section{Design of \toolname} \label{sec:design}

\begin{figure}[t]
    \centering
    \includegraphics[width=\linewidth]{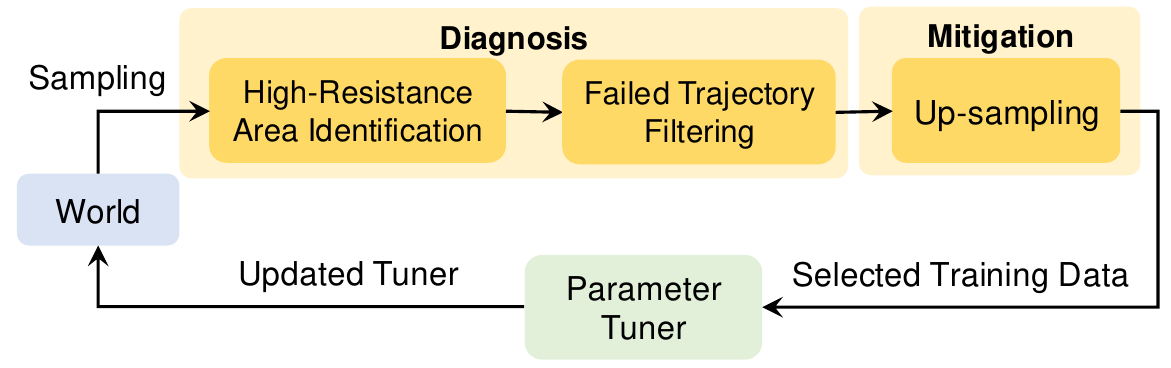}
    \caption{Overview of \toolname Framework}
    \label{fig:overview}
\end{figure}

We introduce \toolname (shown in Fig.~\ref{fig:overview}), with two stages: training inefficiency \emph{diagnosis} (Section~\ref{sec:design:diagnosis}) and \emph{mitigation} (Section~\ref{sec:design:mitigation}) for parameter tuners.

\subsection{Overview}
\label{sec:design:overview}

Alg.~\ref{alg:main} illustrates the integration of \toolname within the parameter tuner training pipeline. Components enhanced by \toolname are highlighted in \textcolor{blue}{\textbf{blue}}, indicating the modifications to the standard parameter tuner. The remaining parts
abstract the conventional components of a deep learning training pipeline.

To better quantitatively pinpoint the \textit{hard-to-pass} areas discussed in Section~\ref{sec:existing}, we develop Alg.~\ref{alg:get_hr_area} and define the concept \textit{high-resistance area} ($\mathcal{H}$), composed of specific poses termed \textit{high-resistance points}. 
Intuitively, \textit{high-resistance} refers to the observation of abrupt \textit{behavioral} changes by the robot at these points, suggesting that these areas pose significant uncertainty to the robot's behavior.

\setlength{\textfloatsep}{10pt}
\begin{algorithm}[t]
\caption{\footnotesize \toolname for Accelerating Parameter Tuner Training}
\label{alg:main}
%\algsetup{linenosize=\footnotesize}
\footnotesize
\begin{algorithmic}[1]

\State {\bfseries Input:} 
Initial parameter tuner $\tau$, high-resistance area $\mathcal{H}$,
environment $\mathcal{E}=f\times \mathcal{W}$ for $\tau$ (where $f$ is classical planner and $\mathcal{W}$ is physical map).

\State {\bfseries Constants:} Robot's initial position ${\tt INIT\_POSE}$, threshold $\lambda$ for sampling initial pose from  $\mathcal{H}$, tuner training iteration $N$, trajectory collection number $K$, tuner update number $L$.

\State {\bfseries Output:} The converged parameter tuner $\tau^{*}$.
\State Set $\mathcal{D}=\textcolor{blue}{\mathcal{P}=\mathcal{H}=\emptyset}$. \Comment{
transition buffer $\mathcal{D}$,
\textcolor{blue}{pose buffer $\mathcal{P}$}
}
\For{iteration $n = 0,...,N$}
\For {$k=0,...,K$}
    \State Initialize state $s_0$ by resetting $\mathcal{E}$. 
    \State Set the robot's initial pose $p_0={\tt INIT\_POSE}$.
    \State Set transition buffer $\mathcal{D}_k=\emptyset$,  \textcolor{blue}{robot pose buffer $\mathcal{P}_k=\emptyset$}.
    \If{\textcolor{blue}{${\tt getRandom}($0,1$)$ < $\lambda$ and $\mathcal{H} \neq \emptyset$}} \label{algo1:line:if}
        \State \textcolor{blue}{$p_0 \leftarrow {\tt getRandom}(\mathcal{H})$ 
        % \Comment{High-resistance area $\mathcal{H}$ up-sampling}.
        } \label{algo1:line:get_random}
    \EndIf
    \State Set $\tilde{s}=[s_0]$ and \textcolor{blue}{$\tilde{p}=[p_0]$}. 
    % \Comment{State trajectory $\tilde{s}$, \textcolor{blue}{pose trajectory $\tilde{p}$}}
    \For{$t=0,...,T$}
        \State Run $\tau$ in $\mathcal{E}$, take tuning adjustment $a_t=\tau(s_t)$.
        \State Obtain feedback $r_t$, state $s_{t+1}$ and \textcolor{blue}{pose $p_{t+1}$}.
        \State $\tilde{s} \leftarrow \tilde{s}.{\tt append}(s_{t+1})$,\ \ \ \textcolor{blue}{$\tilde{p}\leftarrow \tilde{p}.{\tt append}(p_{t+1})$}.
        \If {$\tau$ gets ${\tt done}$ signal from $\mathcal{E}$}, 
            \State $\mathcal{D}_k \leftarrow \mathcal{D}_k \cup \{\tilde{s}\}$, 
        \textcolor{blue}{$\mathcal{P}_k \leftarrow \mathcal{P}_k \cup \{\tilde{p}\}$}.
        \EndIf
    \EndFor
    \State Update $\mathcal{D}$ using $\mathcal{D}_k$. 
    \Comment{This step is algorithm-specific}
    \State \textcolor{blue}{Update $\mathcal{H}$: $\mathcal{H} \leftarrow {\tt getHRArea}(\mathcal{P}_k)$. \Comment{Refer to Alg.~\ref{alg:get_hr_area}.}}\label{algo1:line:gethr}
\EndFor

\For{$\ell=0,...,L$}
    \State $\mathcal{B} \leftarrow {\tt getMiniBatch}(\mathcal{D})$ \Comment{Random mini-batch} 
    \State Update tuner $\tau$ using $\mathcal{B}$. \Comment{Algorithm-specific}
\EndFor
\EndFor
\State Return the final parameter tuner $\tau^*$.
\end{algorithmic}
\end{algorithm}

\subsection{Training Inefficiency Diagnosis} 
\label{sec:design:diagnosis}

\noindent {\bf High-Resistance Point Identification.} 
Alg.~\ref{alg:get_hr_area} identifies high-resistance areas $\mathcal{H}$ within robot's navigation environment. By processing pose trajectories $\mathcal{P}$ and a threshold $\eta$, it outputs $\mathcal{H}$, i.e. poses where the robot exhibits significant behavioral changes, indicating challenging regions for the planner.

The algorithm begins by accepting a set of pose trajectories $\mathcal{P}$. 
For each trajectory in $\mathcal{P}$, the algorithm checks if the trajectory successfully reaches the mission goal, filtering out any mission-failure trajectories (in Line~\ref{algo2:line:filter_out} of Alg.~\ref{alg:get_hr_area}). 
For trajectories that reach the mission goal, the algorithm examines each triplet of consecutive poses $(p_i, p_{i+1}, p_{i+2})$ within the trajectory. The poses are characterized by their position $x, y$ and orientation $w$.

In Line~\ref{algo2:line:rho} of Alg.~\ref{alg:get_hr_area}, the differences in $x$ and $y$ coordinates between consecutive robot poses are used to calculate \textit{trajectory orientations}, notated as $\rho_i=\arctan(\Delta y_{i}/\Delta x_{i})$ and $\rho_{i+1}=\arctan(\Delta y_{i+1}/\Delta x_{i+1})$.
Then variation of trajectory orientations $\Delta \rho_i=\rho_{i+1}-\rho_i$ is then normalized to the range $[-\pi, \pi]$ (in Line~\ref{algo2:line:delta_rho} of Alg.~\ref{alg:get_hr_area}).
If the absolute value $|\Delta \rho_i|$ exceeds the threshold $\eta$, the corresponding pose $p_i$ is identified as a high-resistance point and added to the set $\mathcal{H}$. 
Finally, the algorithm returns the set $\mathcal{H}$, representing the high-resistance area. The key intuition behind Alg.~\ref{alg:get_hr_area} is to capture regions where the parameter tuner
shows high uncertainty in challenging scenarios, indicated by a high variance in the trajectory. \looseness=-1

Note that there are two technical aspects that merit particular attention: 
{\ding{182}}  Alg.~\ref{alg:get_hr_area} employs variations in trajectory orientations $\Delta \rho$ rather than robot orientations $\Delta w$ as the metric for comparison against the threshold $\eta$. This design effectively captures behaviors such as back-and-forth movements, where the robot's orientation may remain relatively unchanged and thus cannot be detected by $\Delta w$. However, these movements are also clear indicators of challenges faced by the planner.
{\ding{183}}  Alg.~\ref{alg:get_hr_area} strategically excludes failure trajectories. This exclusion ensures that any points of high resistance can ultimately reach the goal. Allowing the robot to persist in areas that are excessively challenging would cause the planner to become indefinitely stuck, thereby introducing noise into the training of the parameter tuner.
This process allows for the identification of challenging areas where the robot experiences significant difficulties, enabling targeted interventions to improve the planner's performance in these regions. \looseness=-1

\subsection{Training Inefficiency Mitigation}
\label{sec:design:mitigation}

\noindent \textbf{High-Resistance Area Up-sampling.} In Lines~\ref{algo1:line:if}-\ref{algo1:line:get_random} of Alg.~\ref{alg:main}, the process of high-resistance area up-sampling is implemented to enhance the robustness and adaptability of the parameter tuner during training. 
This algorithm leverages threshold $\lambda \in [0,1]$ to determine whether to sample the robot's initial position from the set of high-resistance areas $\mathcal{H}$.
It ensures that the parameter tuner frequently encounters and learns to navigate through high-resistance regions, thereby improving its ability to handle challenging scenarios. 
By balancing this targeted exploration with a general exploration of the environment, the parameter tuner will train more efficiently and effectively in high-resistance areas.
Additionally, $\mathcal{H}$ is updated iteratively in Line~\ref{algo1:line:gethr} of Alg.~\ref{alg:main}, ensuring that the set of high-resistance areas remains relevant and accurately reflects the current sampling challenges. 
This iterative update is crucial for maintaining the effectiveness of the up-sampling strategy. 
Overall, the up-sampling of high-resistance areas is critical for the parameter tuner's training as it enhances the planner's robustness, promotes better generalization to new environments, and addresses training inefficiency. 

\setlength{\textfloatsep}{10pt}
\begin{algorithm}[t!]
\caption{\footnotesize High-Resistance Area Identification ${\tt getHRArea}$}
\label{alg:get_hr_area}

%\algsetup{linenosize=\tiny}
\footnotesize
\begin{algorithmic}[1]

\State {\bfseries Input: } 
% Initial meta-planner policy network $\tau$, 
Pose trajectory set $\mathcal{P}$,
threshold $\eta$ for high-resistance identification.

% \State {\bfseries Constants:} 
% Robot's initial position ${\tt INIT\_POSE}$, 
% threshold $\lambda$ for sampling initial pose from  $\mathcal{H}$, 
% policy training iteration $N$, 
% trajectory collection number $K$, 
% policy update number $L$.

\State {\bfseries Output:} 
High-resistance area $\mathcal{H}$.

\State Initialize $\mathcal{H}=\emptyset$.

\For{each $\tilde{p}$ in $\mathcal{P}$}
\If{$\tilde{p}$ reaches the mission goal} %\Comment{Filter out mission-failure trajectories} 
\label{algo2:line:filter_out}
\State \Comment{$p$ is a 3-dim vector: position $x, y$ and orientation $w$}
\For{each $(p_{i}, p_{i+1}, p_{i+2})$ in  $\tilde{p}$}  %\Comment{Pick 3 consecutive poses}

    % \State $(x_i, y_i, w_i), (x_{i+1}, y_{i+1}, w_{i+1}), (x_{i+2}, y_{i+2}, w_{i+2}) \leftarrow p_i$, $p_{i+1}, p_{i+2}$.

    \State $(x_k, y_k, w_k) \leftarrow p_k, k=\{i,i+1,i+2\}$, .
    
    \State $\Delta x_i, \Delta x_{i+1} \leftarrow x_{i+1}-x_i, x_{i+2}-x_{i+1}$.

    \State $\Delta y_i, \Delta y_{i+1}\ \leftarrow y_{i+1}-y_i, y_{i+2}-y_{i+1}$.

    \State $\rho_{i}, \rho_{i+1} \leftarrow \arctan\Big(\dfrac{\Delta y_{i}}{\Delta x_{i}}\Big), 
    \arctan\Big(\dfrac{\Delta y_{i+1}}{\Delta x_{i+1}}\Big) $\label{algo2:line:rho}
    % \arctan(\Delta y_{i+1}/ \Delta x_{i+1})$ 
    % \Comment{Trajectory orientations}
    
    \State $\Delta \rho_i \leftarrow \rho_{i+1} - \rho_{i}$.
    % \Comment{Variation of trajectory orientations} 
    \label{algo2:line:delta_rho}
    
    \State $\Delta \rho_i \leftarrow (\Delta \rho_i\mod 2\pi) - \pi$. 
    % \Comment{Normalize $\Delta \rho_i$ to $[-\pi, \pi]$}
    \Comment{Normalization}
    
    \State {{\bf if} $|\Delta \rho_i| > \eta$ {\bf then} $\mathcal{H} \leftarrow \mathcal{H} \cup \{ p_i\}$}  
    % \Comment{Identify $p_i$ as a high-resistance point}
    % \EndIf
\EndFor
\EndIf
\EndFor
\State Return the high-resistance area $\mathcal{H}$.

\end{algorithmic}
\end{algorithm}
\vspace{-6pt}

\section{Evaluation} 
\label{sec:evaluation}

\begin{table*}[t]
  \centering
  \footnotesize
  \tabcolsep=5.2pt
  % \small\addtolength{\tabcolsep}{-3pt}
  % \captionsetup{belowskip=0pt,aboveskip=5pt}
  \caption{Navigation performance of \toolname and baseline methods under \textbf{Same-Env} setup. 
  }
  \label{tab:eval:same_env}
  % \resizebox{\textwidth}{!}{
  \begin{tabular}{cccccccccccccccc}
    \toprule
    \multirow{2}{*}{\makecell[c]{Metric $\Rightarrow$ \\ Tuner $\Downarrow$}} &
    \multicolumn{5}{c}{Easy} &
    \multicolumn{5}{c}{Medium} &
    \multicolumn{5}{c}{Difficult} \\
    \cmidrule(lr){2-6}\cmidrule(lr){7-11}\cmidrule(lr){12-16}
    ~    
    & NS $\uparrow$ & ATT $\downarrow$ & SR $\uparrow$ & CR $\downarrow$ & TR $\downarrow$& NS $\uparrow$ & ATT $\downarrow$ & SR $\uparrow$ & CR $\downarrow$ & TR $\downarrow$& NS $\uparrow$ & ATT $\downarrow$ & SR $\uparrow$ & CR $\downarrow$ & TR $\downarrow$
    \\
    \midrule
    DWA (No Tuner)   & 20.72 & 30.01 & \textbf{97.65} & \textbf{1.18} & \textbf{1.18} 
          & 16.91 & 33.35 & \textbf{90.59} & \textbf{2.35} & 7.06 
          & 10.61 & 30.99 & 53.75 & 16.25 & 30.00 \\
    Manual & 25.44 & 23.60 & 76.47 & 23.53 & 0.00 
             & 16.30 & 30.68 & 64.71 & 23.53 & 11.76 
             & 12.43 & \textbf{27.40} & 50.00 & 18.75 & 31.25 \\
    BayesOpt  & 17.88 &	48.09 &	74.00 &	10.00 &	16.00
              & 15.61 &	21.84 &	49.02  & 21.57 & 29.41	
              & 5.94 &	58.37 &	43.75  & 6.25 &	50.00 \\
    APPLR   & 20.19 & 28.48 & 70.59 & 13.24 &	16.18	
            & 21.36 & 26.36 & 85.29 & 7.35 &	7.35	
            & 12.83 & 33.59 & 57.81 & \textbf{4.69} &	37.50   \\
     Ours & \textbf{35.03} & \textbf{13.45} & 80.88 & 11.76 & 7.35
     & \textbf{23.23} & \textbf{21.62} & 76.74 & 17.65 & \textbf{5.88} 
     & \textbf{14.56} & 27.81 & \textbf{60.94} & 18.75 & \textbf{20.31} \\ 
    \bottomrule
  \end{tabular}
  % }
  \begin{tablenotes}
       \item {\bf NS}: Navigation Score. {\bf ATT}: Actual Traveling Time. {\bf SR}: Success Rate. {\bf CR}: Collision Rate. {\bf TR}: Timeout Rate.
   \end{tablenotes}
\end{table*}

\begin{table*}[t]
  \centering
  \footnotesize
  % \scriptsize
  % \tiny
  \tabcolsep=1.6pt
  % \arraystretch=2.2pt
  \renewcommand{\arraystretch}{1.2}
  % \small\addtolength{\tabcolsep}{-3pt}
  \captionsetup{belowskip=0pt,aboveskip=5pt}
  \caption{
  \small Navigation performance of \toolname and baselines under {\bf Cross-Env} and {\bf Cross-Level} setups. The column names represent the test environment levels: All (across all three levels), Easy, Medium and Difficult.
  % \todo{arrow for metrics. easy different map. bold for best performance. Add levels in the Row.Remove RandSampling.}
  }
  \label{tab:eval:cross_env_level}
  \resizebox{\textwidth}{!}{
  \begin{tabular}{cccccccccccccccccccccc}
  % \begin{tabular}{crrrrrrrrrrrrrrrrrrrr}
    \toprule
    \multirow{2}{*}{\makecell[c]{Training \\ Env.}} & \multirow{2}{*}{\makecell[c]{Metric $\Rightarrow$ \\ Tuner $\Downarrow$}} &
    \multicolumn{5}{c}{All} &
    \multicolumn{5}{c}{Easy} &
    \multicolumn{5}{c}{Medium} &
    \multicolumn{5}{c}{Difficult} \\
    \cmidrule(lr){3-7}\cmidrule(lr){8-12}\cmidrule(lr){13-17}\cmidrule(lr){18-22}
    ~      
    & &  NS $\uparrow$ & ATT $\downarrow$ & SR $\uparrow$ & CR $\downarrow$ & TR $\downarrow$
    & NS $\uparrow$ & ATT $\downarrow$ & SR $\uparrow$ & CR $\downarrow$ & TR $\downarrow$
    & NS $\uparrow$ & ATT $\downarrow$ & SR $\uparrow$ & CR $\downarrow$ & TR $\downarrow$
    & NS $\uparrow$ & ATT $\downarrow$ & SR $\uparrow$ & CR $\downarrow$ & TR $\downarrow$
    \\
    \midrule
    \multirow{3}{*}{Easy} & BayesOpt   &  12.67	& 44.58	& \textbf{72.00}	& \textbf{10.00}	& \textbf{18.00} 
               & 18.33	& 37.00	& \textbf{88.24}	& \textbf{0.00}	& 11.76	
               & 12.68	& 45.81	& \textbf{76.47}	& \textbf{17.65}	& \textbf{5.88}	
               & 6.65	& 56.79	& \textbf{50.00}	& 12.50	& \textbf{37.50} \\
    & APPLR      &  10.46 & 37.91 &	47.60 & 19.60 & 32.80 
               & 20.04 &  29.82 &	71.76 &	12.94 & 15.29 
               &	7.06 & 44.63 & 44.71 & \textbf{17.65} & 37.65 
               & 3.90 & 49.80 & 25.00 & 28.75 & 46.25 \\
    % RandSampling & 14.10 & 22.22  & 54.80 & 10.80 & 32.40 
    %              &  24.92 &	15.66 &	80.00 & 10.00 &	10.00 
    %              & 11.21 & 22.40 & 49.41 & 14.12 & 36.47 
    %              & 5.67 & 33.47 & 33.75 & 18.75 & 47.50 \\
    & Ours      & \textbf{20.43} & \textbf{19.04} &	58.40 & 15.20 & 26.40  
              & \textbf{35.32} & \textbf{14.08} &	82.35 &	11.76 & \textbf{5.88}
              & \textbf{15.26 } & \textbf{22.16} & 50.59 & 22.35 & 27.06 
              & \textbf{10.10} & \textbf{25.48} & 41.25 & \textbf{11.25} & 47.50\\
    
    \midrule
    \multirow{3}{*}{Medium} & BayesOpt    & 19.74	& \textbf{17.68} &	52.00 &	16.00 &	32.00 
                & 35.18 & \textbf{10.78} & 	76.47 &\textbf{	5.88} &	17.65	
                & 13.38 & 25.33 & 	41.18 &	29.41 &	29.41	
                & 10.09 & \textbf{23.72} & 	37.50 &	\textbf{12.50} &	50.00\\
    & APPLR       & 21.85 & 25.39 &	\textbf{77.60} & \textbf{11.60} &	11.80 
                & 31.43 &	20.71 &	90.59 &	9.41 &	\textbf{0.00 }
                & 21.49 &	26.89 &	\textbf{84.71} &	\textbf{7.06} &	8.24 
                & 12.06 &	31.02 &	\textbf{56.25} &	18.75 &	\textbf{25.00}  \\
    % RandSampling & 20.07 & 23.71 &	66.00 &	14.40 &	19.60  
    %              & 32.90 &	17.92 &	89.41 &	9.41 &	1.18 
    %              & 18.84 &	25.04 &	68.24 &	17.65 &	14.12 
    %              & 7.75 &	35.39 &	38.75 &	16.25 &	45.00\\
    & Ours     &\textbf{ 24.39} & 22.22 & 75.20 & 13.60 & \textbf{11.20}  
             & \textbf{37.14} & 15.66 & \textbf{92.94} & 7.06 & \textbf{0.00} 
             & \textbf{22.61} & \textbf{22.40} & 75.29 & 18.82 & \textbf{5.88 }
             & \textbf{12.73} & 33.47 & \textbf{56.25} & 15.00 & 28.75\\
    
    \midrule
    \multirow{3}{*}{Difficult} & BayesOpt   & 8.72  &  52.23 &  52.00 & 10.00 & 38.00 
               & 11.32 &  43.11 &  58.82 &	5.88  &	35.29	
               & 8.87  &  54.66 &  52.94 &	\textbf{5.88}  &	41.18	
               & 5.81  &  62.13 &  43.75 &	18.75 &	37.50 \\
    & APPLR       & 22.56 & 25.21 &	76.00 &	\textbf{8.40}  &	15.60 
                & 34.51 & 18.93 &	92.94 &	5.88  &	1.18 
                & 19.65 & 26.69 &	75.29 &	12.94 &	11.76 
                & 12.97 & 33.74 &	58.75 &	\textbf{6.25}  &	35.00  \\
    % RandSampling &  20.24 &  27.17 & 73.60 & 7.20 &	19.20 
    %              &  31.68 &	19.05 &	90.59 &	5.88 &	3.53 
    %              &  16.73 &	32.74 &	72.94 &	8.24 &	18.82 
    %              &  11.81 &	33.38 &	56.25 &	7.50 &	36.25\\
    & Ours    & \textbf{24.61} & \textbf{22.54} & \textbf{79.60} &	12.00 &	8.40 
            & \textbf{36.30 } & \textbf{17.90} & \textbf{96.47} &	\textbf{3.53}  &	\textbf{0.00 }
            & \textbf{22.16} & \textbf{24.53} & \textbf{80.00} &	12.94 &	\textbf{7.06 }
            & \textbf{14.78} &\textbf{ 27.56} & \textbf{61.25} &	20.00 &	\textbf{18.75}\\
    \bottomrule
  \end{tabular}
  }
  \begin{tablenotes}
  \scriptsize
       \item {\bf NS}: Navigation Score. {\bf ATT}: Actual Traveling Time. {\bf SR}: Success Rate. {\bf CR}: Collision Rate. {\bf TR}: Timeout Rate.
       % \item [2] Some UTF-8 support via \LaTeX{} kernel commands.
   \end{tablenotes}
   \vspace{-10pt}
\end{table*}

\noindent In evaluation, we answer research questions ({\bf RQ}) below: \\
{\bf RQ1}: How is \toolname's tuning performance compared with baseline tuning methods (under the same time budget)? \\
{\bf RQ2}: Can \toolname improve parameter tuner's robustness under different environments and difficulty levels? \\
{\bf RQ3:} How effectiveness is \toolname's diagnosis (region identification and failure filtering) for parameter tuners? \\
{\bf RQ4:} Can \toolname improve the efficiency of training process of parameter tuners? \\
{\bf RQ5:} Can the advantage of \toolname be illustrated through a case study?

\subsection{Experiment Setup} 
\label{sec:expr_setup}

\noindent {\bf Simulation \& Planners.} Following existing works~\cite{xiao2020appld, xu2021applr}, we implement \toolname using the widely-used Gazebo~\cite{Gazebo} simulator with a ClearPath Jackal differential drive ground robot as a robot model. 
Due to space limit, we leave the details of the classical planner and the parameter tuner that \toolname builds upon for evaluation to Appendix~\ref{sec:appx:exper_setup}.

\noindent {\bf Baselines \& Dataset.} 
We compare \toolname with the following tuning setups:
(1) classical planners (DWA~\cite{fox1997dwa} without parameter tuner; 
(2) manually-tuned by expert (DWA-Fast~\cite{dwafast});
(3) one representative static parameter tuner, BayesOpt~\cite{snoek2012practical} and 
(4) one SOTA dynamic parameter tuner APPLR~\cite{xu2021applr}.
% \todosw{what are the other baselines...}
% \todosw{what are the configuration space..}
We choose the same dataset as APPLR, namely BARN~\cite{perille2020benchmarking} dataset, as it includes various maps for robot navigation inside the Gazebo simulator and it has different difficulty levels' map (i.e., Easy, Medium, and Difficult) to evaluate.
This allows for a convenient evaluation of various planners' performance across these varied settings.

\noindent {\bf Metrics.} We evaluate navigation systems using five metrics:
{\ding{182}}  Actual Traveling Time ({\bf ATT}): Time (in seconds) to complete successful missions.
{\ding{183}}   Mission Success Rate ({\bf SR}): Percentage of missions completed (i.e., reaching the goal without collisions and within the time budget.).
{\ding{184}}  Collision Rate ({\bf CR}): Percentage of missions failed due to collisions with obstacles.
{\ding{185}}  Timeout Rate ({\bf TR}): Percentage of missions where the robot failed to reach the goal within the time limit.
{\ding{186}}  Navigation Score ({\bf NS}): A comprehensive performance metric from BARN Challenge~\cite{barn_challenge}, calculated as follows:
\begin{equation}
\begin{aligned}
    % \footnotesize
    \small
    {\rm {\bf NS}_i} =  \mathbb{I}({\text{Success}}) \times \dfrac{OT_i}{{\rm clip}({\bf ATT}_i, 2OT_{i}, 8OT_{i})}
    \label{eq:ns_score}
\end{aligned}
\end{equation}

Here, $OT_i$ represents the optimal time (computed using Dijkstra algorithm~\cite{dijkstra2022note} in advance), and ${\rm clip}({\bf ATT}, 2OT_{i}, 8OT_{i})$ ensures that the actual traveling time is constrained within twice and eight times the optimal time. 
$\mathbb{I}(\textrm{Success})$ is an indicator function that denotes whether the current navigation mission is successful, consistent with the definition of SR.
Higher values of NS and SR indicate better performance, while lower values of ATT, CR, and TR indicate better performance.

\noindent {\bf Comparison Setup.} We introduce 3 evaluation setups: (1) {\bf Same-Env}: The parameter tuner is trained and tested in the same environment set with predefined but different start positions to avoid overfitting.
(2) {\bf Cross-Env}: The parameter tuner is trained and tested at the same difficulty level but in different environments. 
(3) {\bf Cross-Level}: The parameter tuner is trained on one difficulty level but tested on a different difficulty level.

\subsection{Same-environment Performance (\textbf{RQ1}) }
\label{sec:eval:same_env}

We evaluate \toolname and four baselines under the {\bf Same-Env} setup. DWA, DWA-Fast, and BayesOpt are static planners, so we directly implement their fixed configurations from the training environment and test them on the same environments with random starting positions. For APPLR and \toolname, we train the parameter tuner on the training set and deploy it on the test environments. \looseness=-1

Table~\ref{tab:eval:same_env} shows that \toolname consistently achieves the highest NS score across all difficulty levels, demonstrating its ability to identify and improve performance in underrepresented scenarios. 
While DWA performs better on SR and CR in easy and medium environments, \toolname excels in the comprehensive NS metric. DWA's conservative design prioritizes safety over time efficiency, leading to much longer navigation times, as seen in reduced ATT and lower NS scores, making it less practical and ideal for navigation tasks. Since NS is a widely-used metric~\cite{xu2021applr, xiao2020appld}, \toolname's higher NS score effectively demonstrates its superiority over both static planners and the SOTA dynamic parameter tuner.

\begin{figure*}[t!]
    \centering
    \begin{subfigure}[b]{0.3\linewidth}
        \centering
        \includegraphics[width=1.1\textwidth]{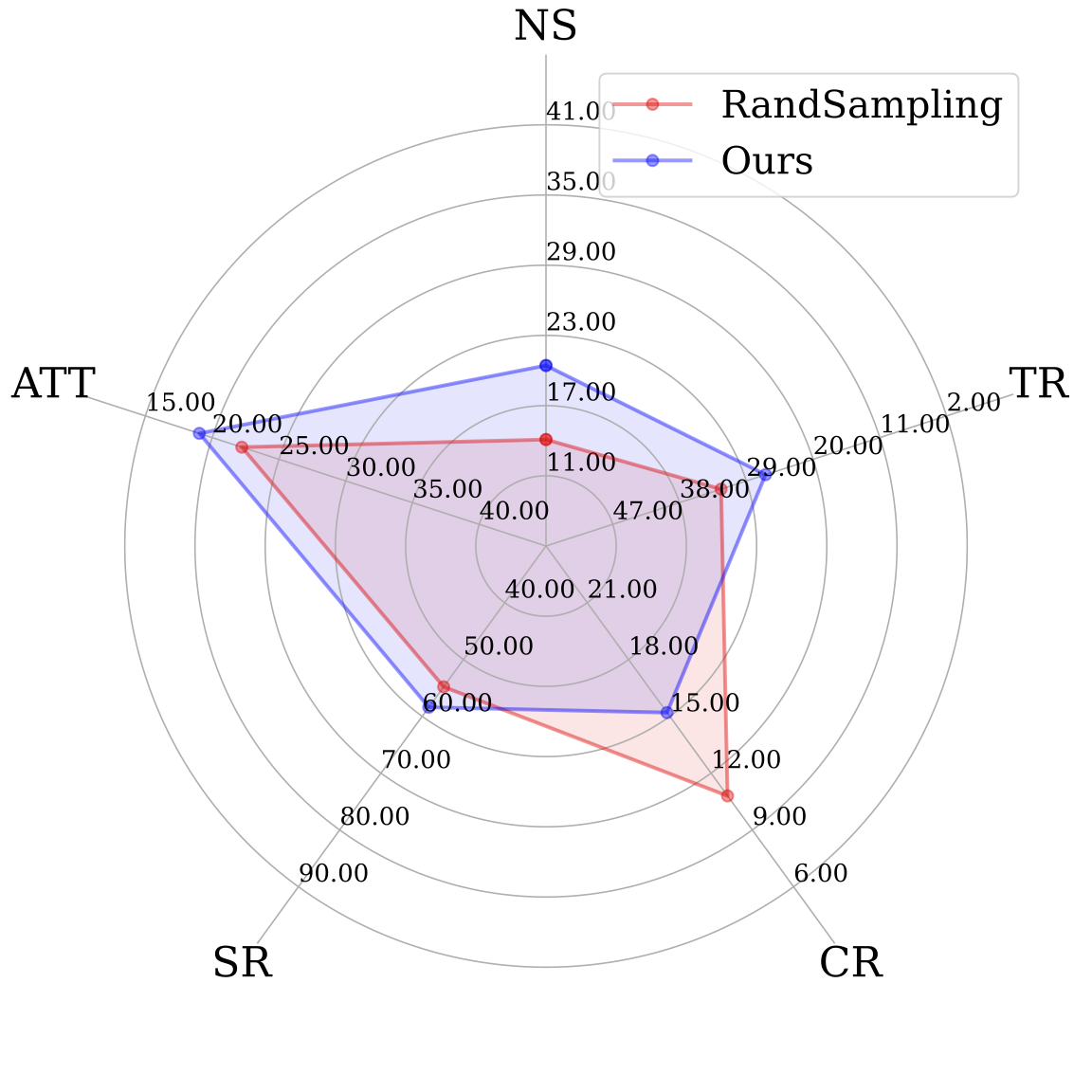}
        \caption{Trained on Easy Level}%\label{fig:}
    \end{subfigure}
    \hspace{9pt}
    \begin{subfigure}[b]{0.3\linewidth}
        \centering
        \includegraphics[width=1.1\textwidth]{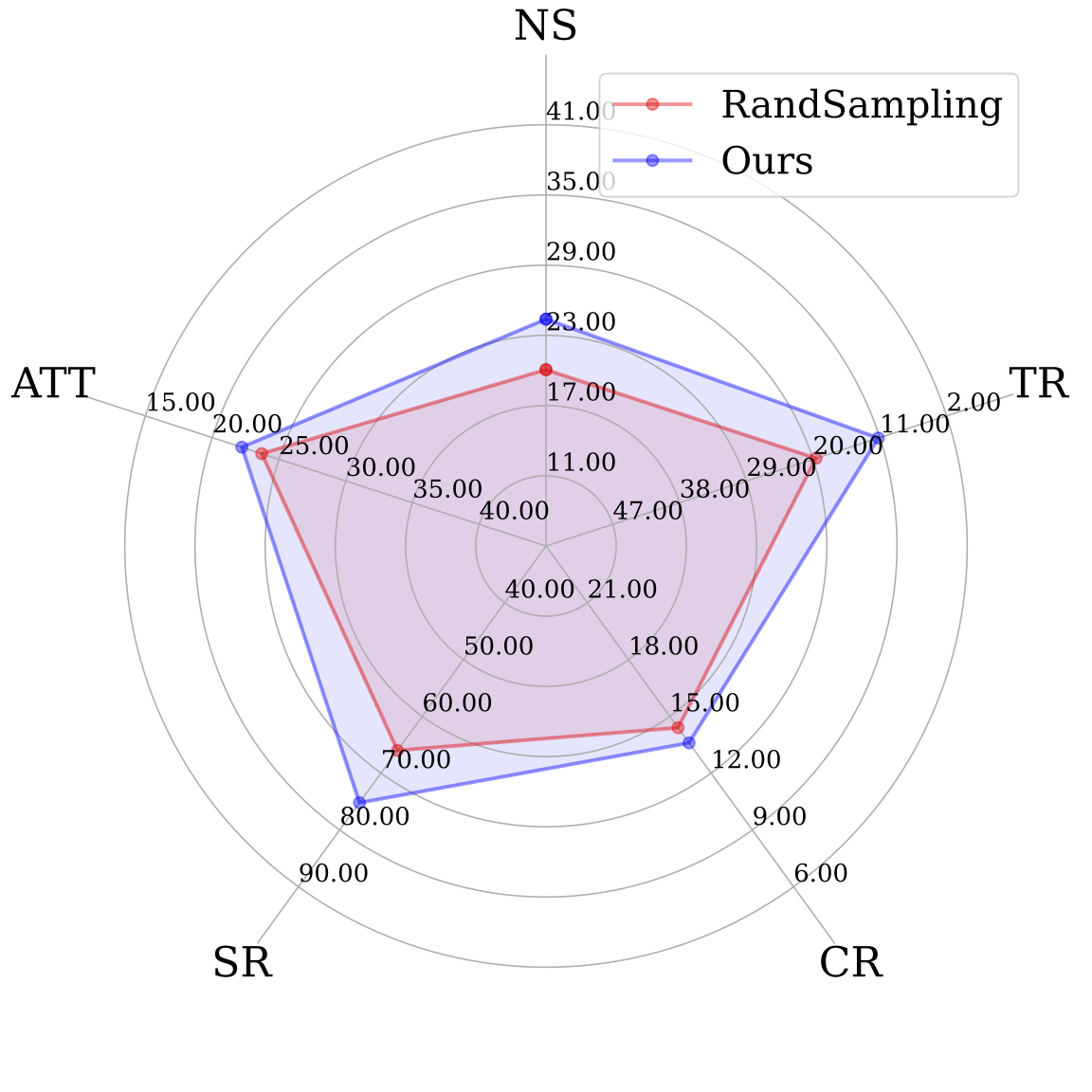}
        \caption{Trained on Medium Level}%\label{fig:}
    \end{subfigure}
    \hspace{9pt}
    \begin{subfigure}[b]{0.3\linewidth}
        \centering
        \includegraphics[width=1.1\textwidth]{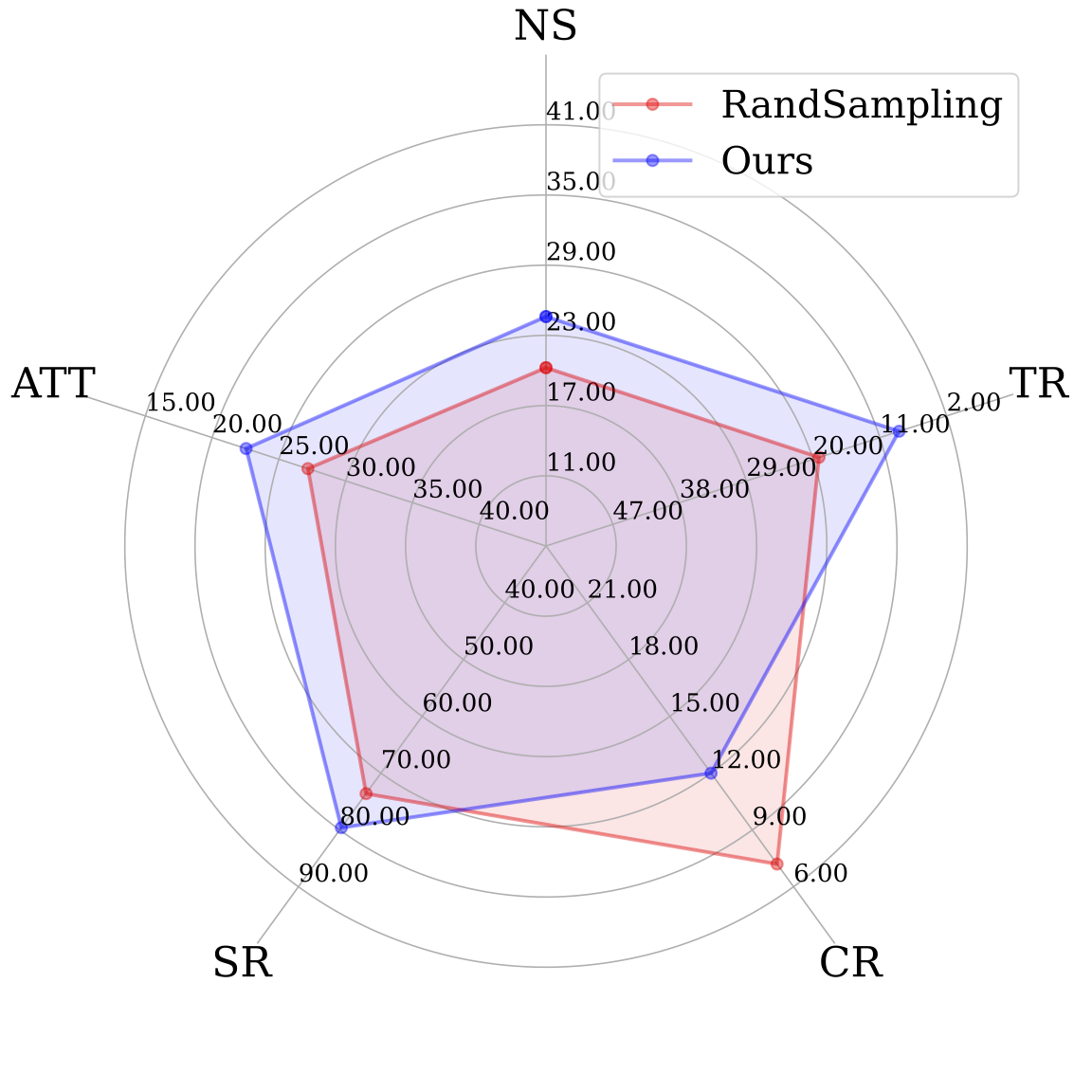}
        \caption{Trained on Difficult Level}%\label{fig:}
    \end{subfigure}
    \caption{\small Comparison of RandomSampling and \toolname. Parameter tuners are tested on all 3 levels.}
    \label{fig:rand_all}
\end{figure*}

\subsection{Cross-environment \& Cross-level Performance (\textbf{RQ2})} 
\label{sec:eval:cross_env}

In this section, we compare \toolname and selected baselines under the \textbf{Cross-Env} and \textbf{Cross-Level} setup, as shown in the diagonal sub-tables in Table~\ref{tab:eval:cross_env_level}.
Under the \textbf{Cross-Env} setup, \toolname consistently achieves the highest NS score across all three environments, indicating its effectiveness in identifying and addressing underrepresented scenarios to enhance planner performance.
For the \textbf{Cross-Level} experiments, \toolname excels in generalization by achieving the highest NS scores when the trained parameter tuner is tested across different environmental difficulty levels. 
This demonstrates \toolname's robustness and capability to maintain navigation efficacy in new and varying conditions. 
Additionally, when parameter tuners trained in difficult environments are applied to easier ones, \toolname outperforms even those baselines specifically trained for simpler environments, while this trend is not observed in the baseline methods.
This result suggests that the parameter tuners trained by our method can learn more sophisticated and adaptable configurations, enabling \toolname to excel in environments that are less challenging.

\subsection{Diagnosis Effectiveness (\textbf{RQ3}) }
\label{sec:eval:diagnosis}

To better understand why our diagnosis strategy is effective, we introduce a variant called RandomSampling (RS).
Unlike \toolname, which identifies high-resistance areas with Alg.~\ref{alg:get_hr_area}, RS indiscriminately samples from all locations reached by the robot during training without strategic guidance, while all other designs remain consistent with \toolname. 
Results in Fig.~\ref{fig:rand_all} show that \toolname significantly outperforms RS, with higher \textrm{NS} scores in all environments. 
The reasons are two-fold.
First, RS fails to expose the robot to challenging, hard-to-reach scenarios, limiting effective training in these high-resistance areas, which is a crucial aspect for satisfactory navigation performance.
Second, as shown in Fig.~\ref{fig:gazebo_3d}, medium and hard environments have more complicated maps with many hard-to-reach scenarios and obstacles than the easy level.
Random sampling in such environments often leads to the robot encountering ``dead-corner'' scenarios -- areas that are hard to exit, leading to extended, unrewarding training trajectories.
Such trajectories will destabilize the training of parameter tuner and degrade navigation performance.
This analysis verifies the necessity of our high-resistance area up-sampling step in \toolname.
By strategically selecting those hard-to-reach areas, we enhance the robot's ability to perform well across a variety of navigational scenarios.

We also conduct an ablation study about failure trajectory filtering.
Table~\ref{tab:eval:failure_filtering} shows the impact of applying failure filtering on \toolname's performance metrics within the same computational budget. With filtering enabled, the Navigation Score (NS) improved to 23.23, demonstrating better navigation quality. Additionally, the Success Rate (SR) increased significantly from 70.59\% to 76.47\%, accompanied by a notable reduction in the Collision Rate (CR) from 22.06\% to 17.65\%. 
This indicates that the filtering process effectively removes noisy or less beneficial training data, enabling the parameter tuner to train more efficiently and enhancing overall robot navigation performance.
Appendix~\ref{sec:appx:diagnosis_effectiveness} shows more results in the comparison between
RS and \toolname.

\begin{figure}[t]
    \centering
    \includegraphics[width=0.7\linewidth]{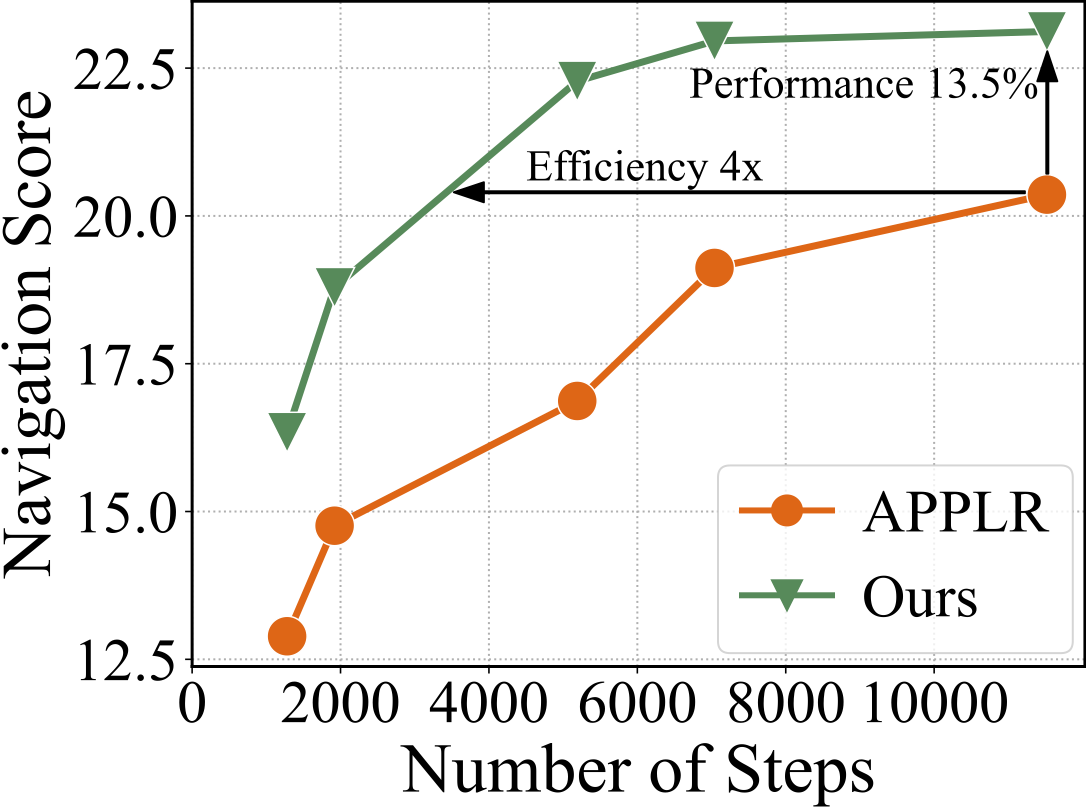}
    \caption{Navigation score ({\bf NS}) of different checkpoints.}
\label{fig:eval:nav_curve}
\end{figure}

\subsection{Training Efficiency (\textbf{RQ4})}
\label{sec:eval:efficiency}

In this section, we evaluate the performance of different parameter tuner checkpoints during training under the \textbf{Cross-Env} setup when \toolname is applied in a medium-level environment.
Fig.~\ref{fig:eval:nav_curve} shows the average \textrm{NS} metric when those checkpoints are evaluated in testing environments.
\toolname consistently outperforms APPLR, achieving higher navigation scores at all training stages with the same training steps.
Furthermore, APPLR requires over \emph{four times} training steps to achieve comparable performance of \toolname, and our method ultimately outperforms APPLR by 13.5\%.
These results demonstrate that our method significantly enhances the training efficiency of the parameter tuner and improves the overall navigation performance of the robot.
Besides NS, we also show more metrics of different parameter tuner checkpoints during training (Appendix~\ref{sec:appx:efficiency}).

\begin{table}[t]
    \centering
    \caption{Ablation Study. Performance of \toolname with and without applying failure filtering.}
    \begin{tabular}{lccccc}
        \toprule
             & wo/ Filtering & w/ Filtering \\
        \midrule
        NS $\uparrow$ & 22.22 & \textbf{23.23} \\
        ATT $\downarrow$ & \textbf{19.60} & 21.62 \\
        SR $\uparrow$ & 70.59  & \textbf{76.47}\\
        CR $\downarrow$ & 22.06  & \textbf{17.65}\\
        TR $\downarrow$ & 7.35 &  \textbf{5.88}  \\
        \bottomrule
        \end{tabular}
    \label{tab:eval:failure_filtering}
\end{table}

\subsection{Case Study (\textbf{RQ5})} \label{sec:eval:case_study}
\begin{figure}[t]
    \centering
    \begin{subfigure}[t]{.48\linewidth}
        \centering
        \includegraphics[width=1\textwidth]{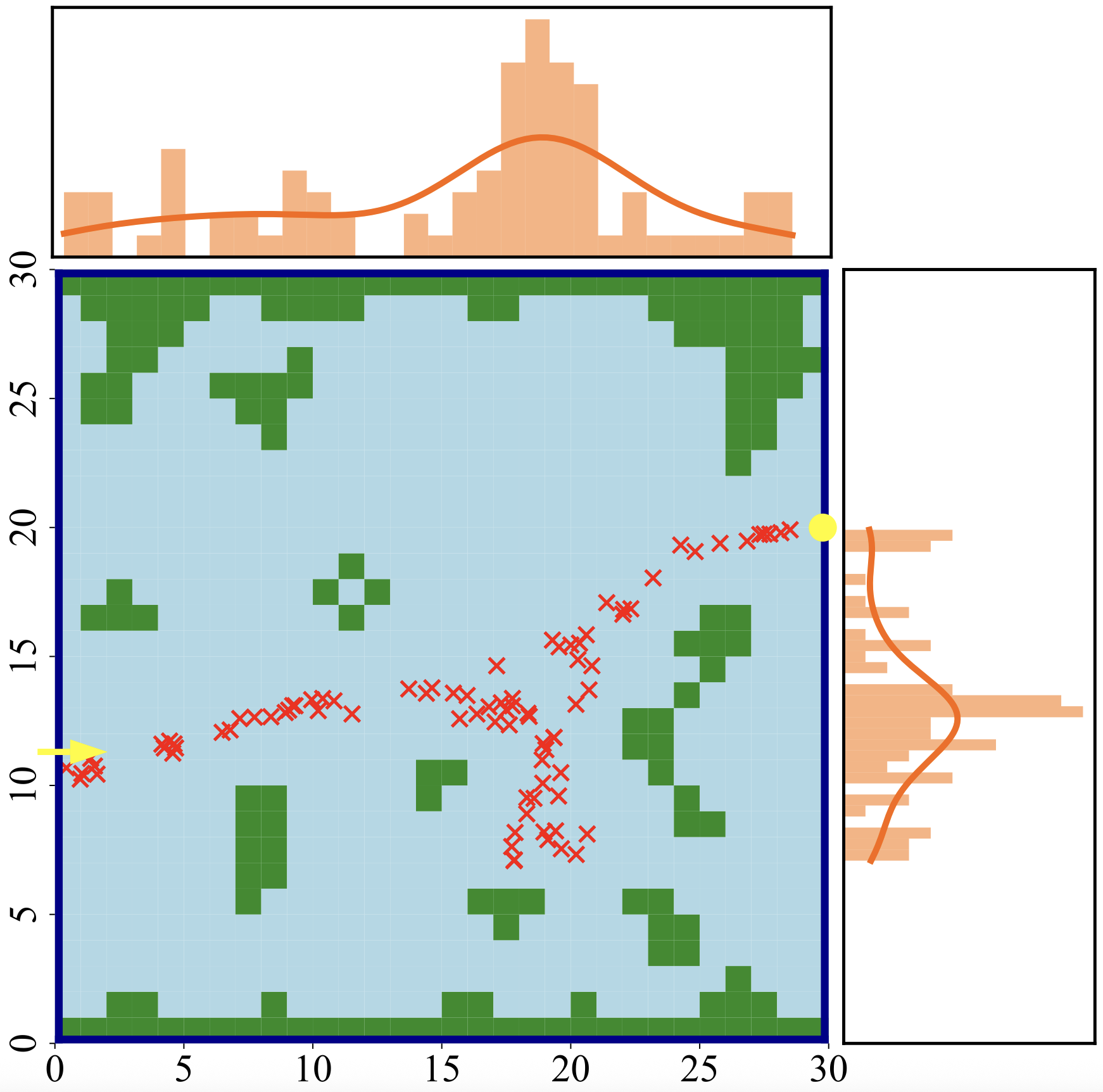}
        \caption{Original: Iter. 10 (Fig.~\ref{fig:moti_b})}
        \label{fig:eval_case_10_before}
    \end{subfigure}
    \begin{subfigure}[t]{.48\linewidth}
        \centering
        \includegraphics[width=1\textwidth]{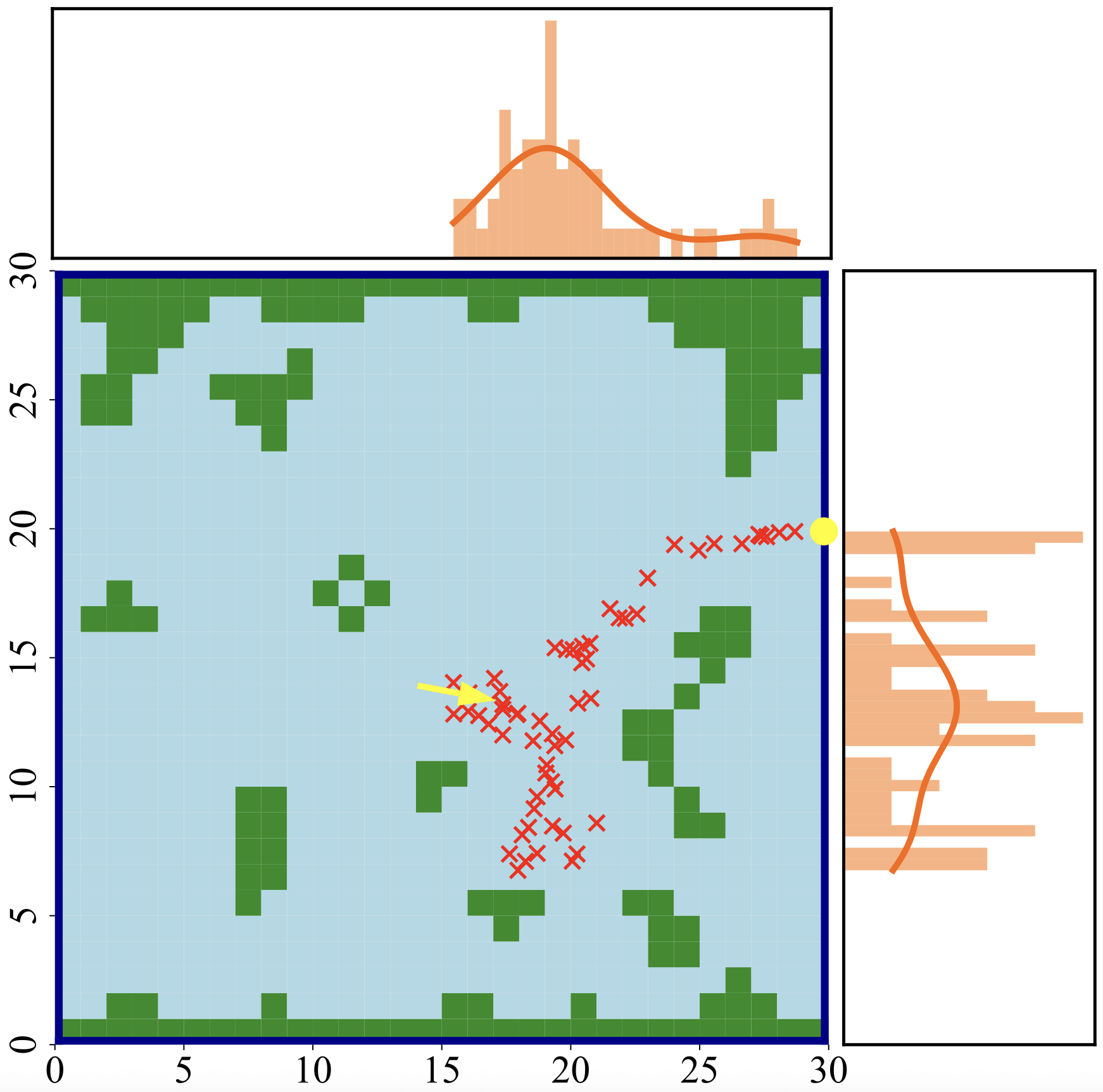}
        \caption{With \toolname: Iter. 10}
        \label{fig:eval_case_10_after}
    \end{subfigure}
    \begin{subfigure}[t]{.48\linewidth}
        \centering
        \includegraphics[width=1\textwidth]{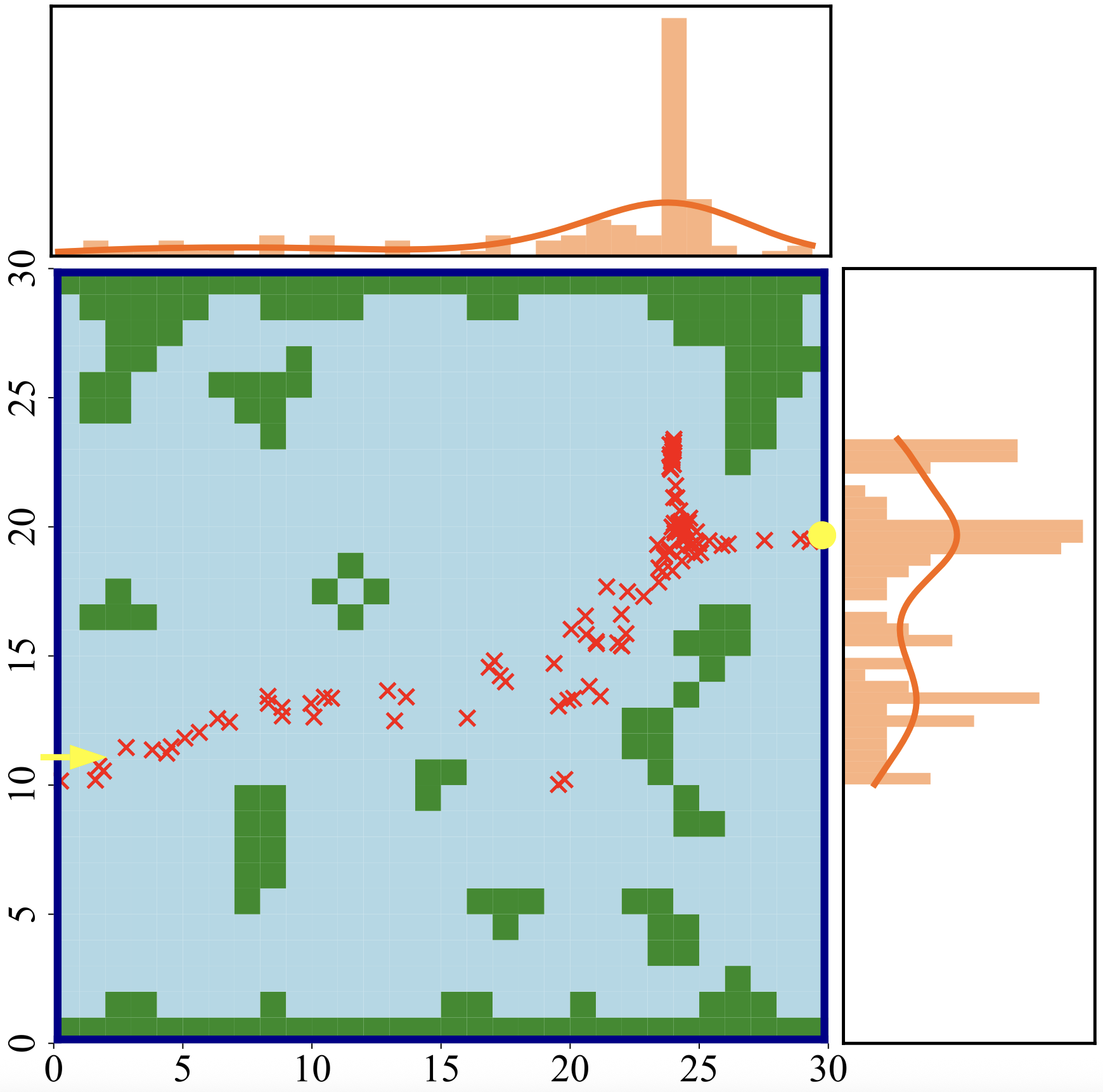}
        \caption{Original: Iter. 90 (Fig.~\ref{fig:moti_d})}
        \label{fig:eval_case_90_before}
    \end{subfigure}
    \begin{subfigure}[t]{.48\linewidth}
        \centering
        \includegraphics[width=1\textwidth]{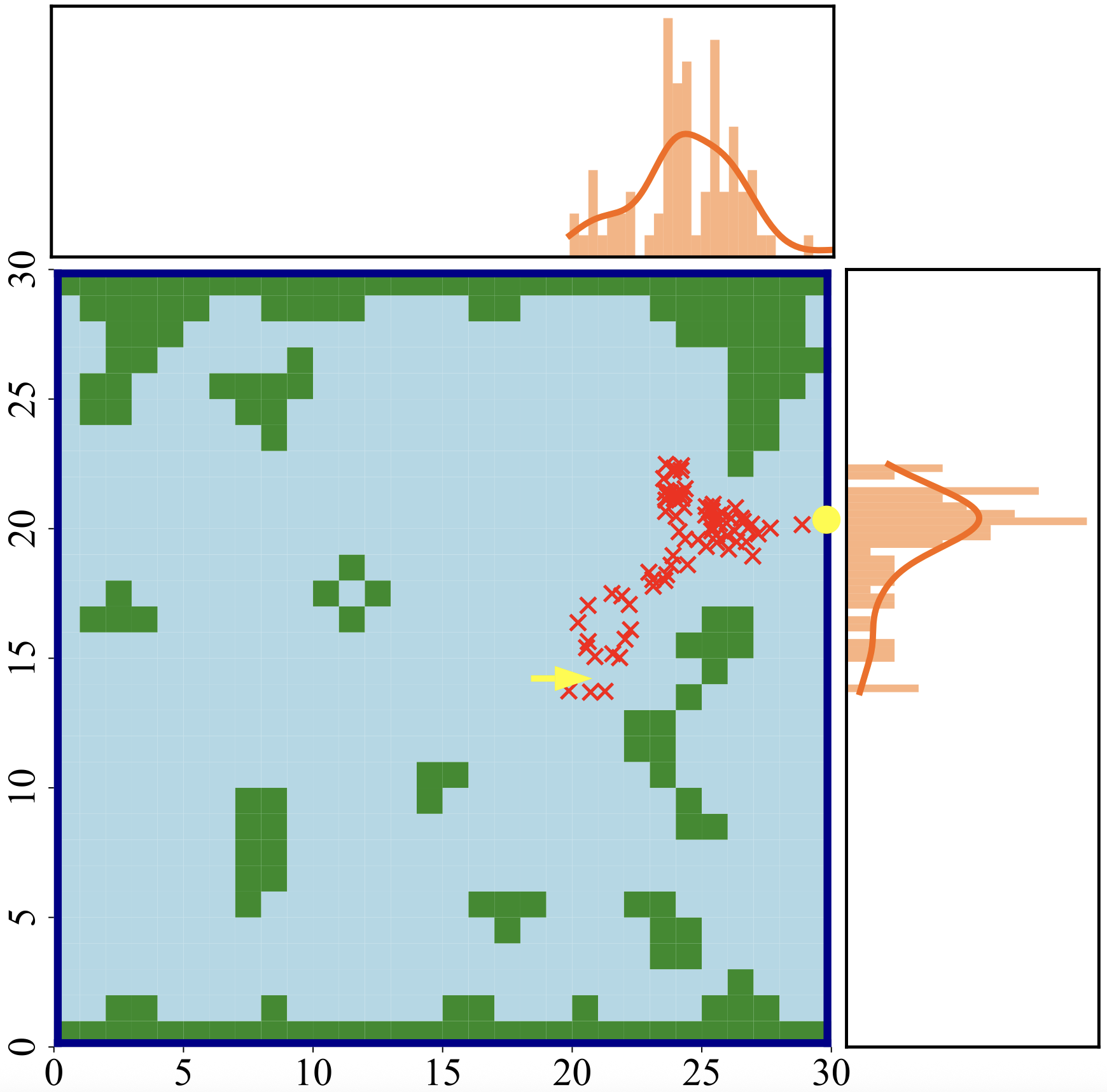}
        \caption{With \toolname: Iter. 90}
        \label{fig:eval_case_90_after}
    \end{subfigure}
    \caption{Case Study for Illustration. The \textbf{\color{Goldenrod}{yellow arrow}} denotes the start position and the \textbf{\color{Goldenrod}{yellow circle}} denotes the goal. }
    \label{fig:eval_case}
\end{figure}

In Fig.~\ref{fig:eval_case}, we show a comparison before and after applying \toolname to the parameter tuner. 
By leveraging our diagnosis and up-sampling techniques, the training process no longer always begins from the default position. Instead, it starts directly in the critical region, where the robot behavior exhibits signals of struggle and challenge. By skipping less beneficial training data and emphasizing the critical samples, we significantly boost the training efficiency of the parameter tuners. \looseness=-1

\subsection{Ablation study}
We conduct ablation study to justify the importance of key components in \toolname and the sensitivity test against the key hyper-parameters. Details can be found in Appendix~\ref{sec:appx:ablation}.

\section{Conclusion} 
\label{sec:conclusion}

In this paper, we identify and analyze the training inefficiency of parameter tuners in robot navigation systems. 
To address these issues, we propose \toolname, a diagnostic and mitigation framework for parameter tuners. \toolname analyzes robot behaviors, identifies exploration bottlenecks and applies a targeted up-sampling strategy to balance the sampling data, enabling more effective learning and exploration. Our experiments demonstrate that \toolname improves navigation performance by 13.5\%+, enhances robustness in unseen environments, and quadruples training efficiency. \looseness=-1

\noindent {\bf Limitations.} 
Although \toolname greatly boosts efficiency, further optimization of training duration is possible. The current pipeline depends on executing the classical planner to collect data, meaning increased complexity of the classical planner can raise training costs.

\section*{Acknowledgement}
We are grateful to the Center for AI Safety for providing computational resources. This work was funded in part by the National Science Foundation (NSF) Awards SHF-1901242, SHF-1910300, Proto-OKN 2333736, IIS-2416835, DARPA VSPELLS - HR001120S0058, ONR N00014-23-1-2081, and Amazon. Any opinions, findings and conclusions or recommendations expressed in this material are those of the authors and do not necessarily reflect the views of the sponsors. \looseness=-1

% \newpage
\bibliographystyle{IEEEtran}
\bibliography{main}

\begin{table*}[t]
    \footnotesize
    \centering
    \resizebox{\textwidth}{!}{\begin{tabular}{cccc}
    \toprule
       Configuration  & Min & Max & Meaning \\
    \midrule
       ${\tt max\_vel\_x}$ & 0.1 & 2.0 & m/s. Maximum threshold for linear velocity. \\
       ${\tt max\_vel\_theta}$ & 0.314 & 3.14 & rad/s. The absolute value of the maximum rotational velocity for the robot. \\
       ${\tt vx\_samples}$     & 4 & 12 & The number of samples to use when exploring the x velocity space. \\
       ${\tt vtheta\_samples}$ & 8 & 40 & The number of samples to use when exploring the theta velocity space\\
       ${\tt path\_distance\_bias}$ & 0.1 & 0.5 & The weight for how much the robot should stay close to the path it was given.  \\
       ${\tt goal\_distance\_bias}$ & 0.1 & 2 & 
       The weight for how much the robot should attempt to reach its local goal.\\
       ${\tt inflation\_radius}$  & 0.1 & 0.6 & Controls how far away the zero cost point is from obstacle.\\
    \midrule
    \end{tabular}}
    \caption{Action Space of Parameter Tuner Policy $\pi$.}
    \label{tab:appx:action_space}
\end{table*}

\setcounter{section}{0}
\newpage
\section*{Appendix}
\section{Experiment Setup} \label{sec:appx:exper_setup}

We run experiments on Ubuntu 20.04, with Intel i9-13900K, 64 GB RAM and Nvidia GPU RTX 2080.
The robot is equipped with a 720-dimensional planar laser scanner with a 270$^{\circ}$ field of view, which provides our sensory input $o_t$. 
We pre-process the LiDAR scans by capping the maximum range to 2m.
The onboard Robot Operating System (ROS) ${\tt move\_base}$ navigation stack (our underlying classical motion planner $f$) uses Dijkstra’s algorithm~\cite{dijkstra2022note} to plan a global path and runs DWA~\cite{fox1997dwa} as the local classical planner to follow the global path.

The parameter tuner policy $\pi$ is trained to update the planning configurations $\theta$ of the classical planner. 
We maintain the same settings as APPLR, utilizing the DWA classical planner and the same action space $\theta$, including ${\tt max\_vel\_x}$, ${\tt max\_vel\_theta}$, ${\tt vx\_samples}$, ${\tt vtheta\_samples}$, ${\tt occdist\_scale}$, ${\tt pdist\_scale}$, ${\tt gdist\_scale}$, and ${\tt inflation\_radius}$.\footnote{https://wiki.ros.org/dwa\_local\_planner}
We use the ROS dynamic reconfigure client to dynamically change planner parameters. 
The global goal $\alpha$ is assigned manually, while $\theta_0$ is the default set of parameters provided by the robot manufacturer.\footnote{http://wiki.ros.org/costmap\_2d}
Table~\ref{tab:appx:action_space} shows the details of each configuration of the parameter tuner.\\

\section{Diagnosis Effectiveness}
\label{sec:appx:diagnosis_effectiveness}

\begin{figure*}[h]
    \centering
    \begin{subfigure}[b]{0.325\linewidth}
        \centering
        \includegraphics[width=\textwidth]{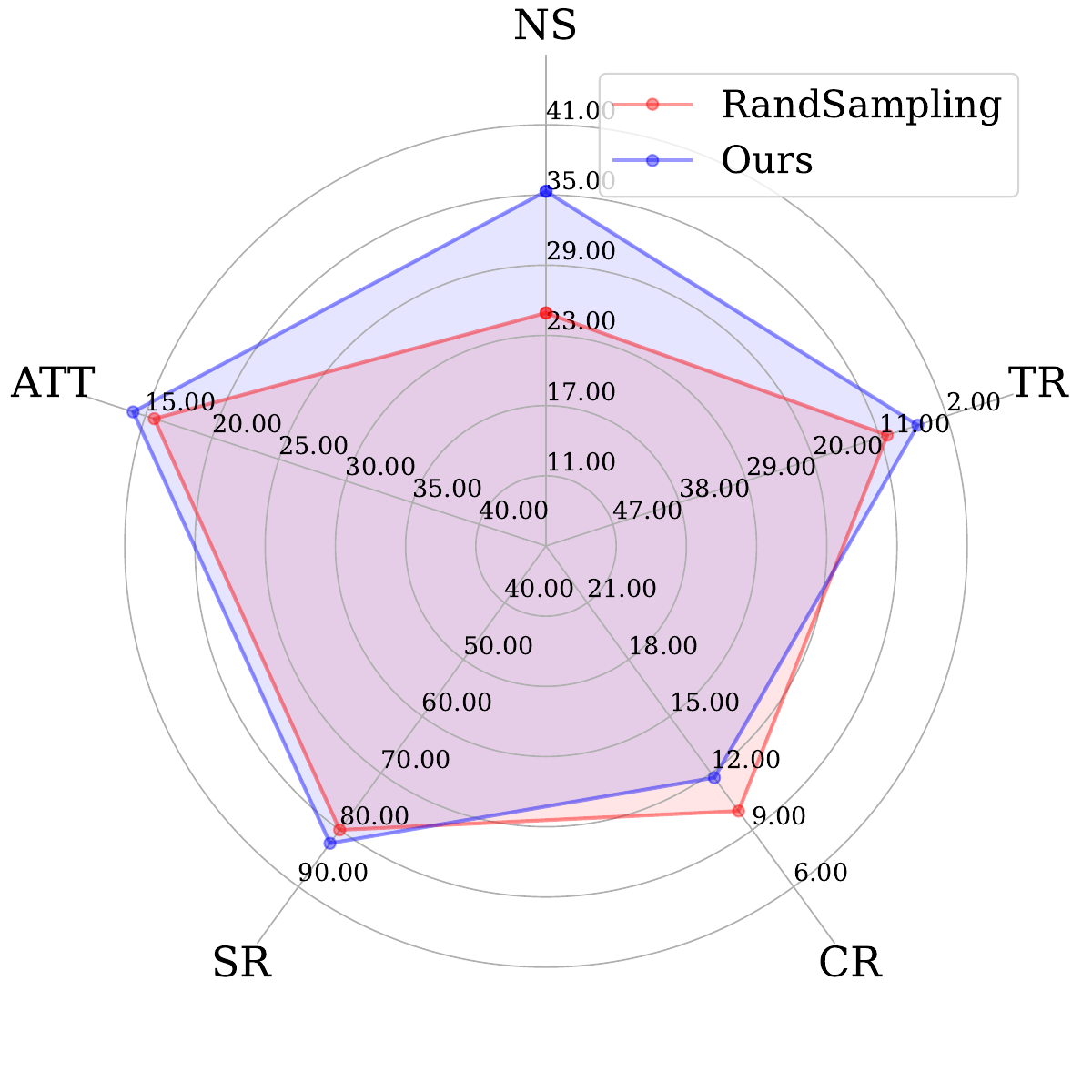}
        \caption{Train on Easy} %\label{fig:}
    \end{subfigure}
    \begin{subfigure}[b]{0.325\linewidth}
        \centering
        \includegraphics[width=\textwidth]{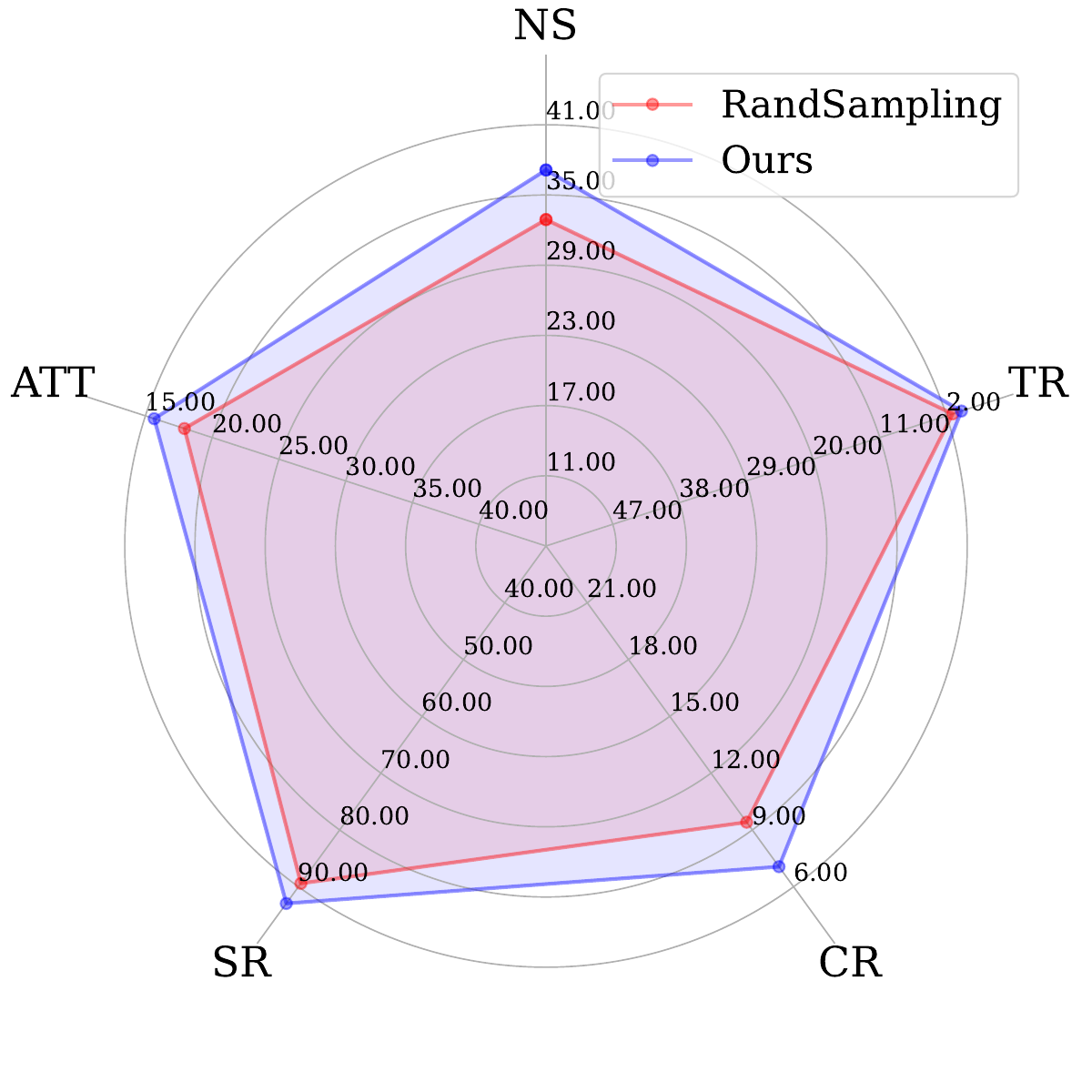}
        \caption{Train on Medium} %\label{fig:}
    \end{subfigure}
    \begin{subfigure}[b]{0.325\linewidth}
        \centering
        \includegraphics[width=\textwidth]{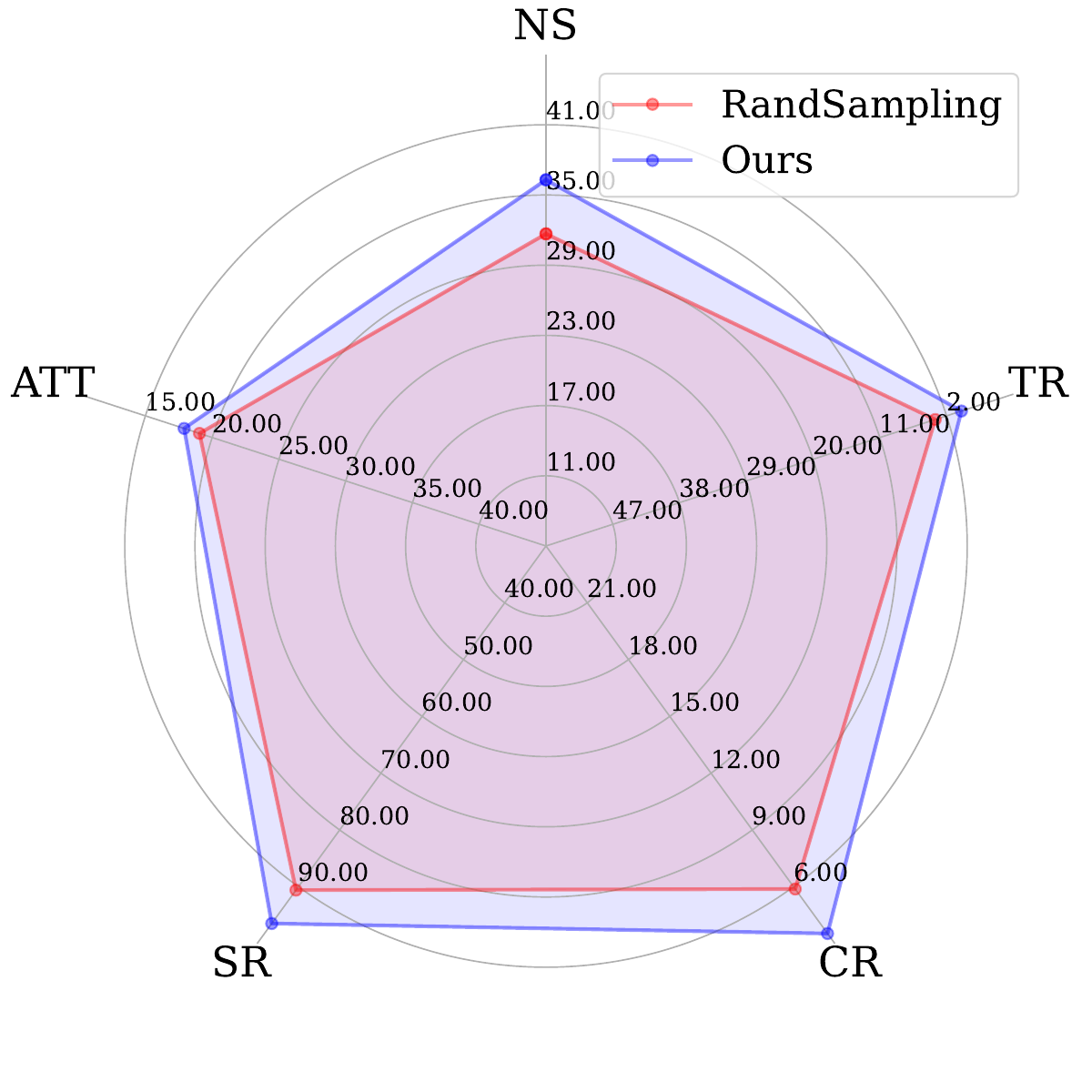}
        \caption{Train on Difficult} %\label{fig:}
    \end{subfigure}
    \caption{RandSampling v.s. \toolname. Test on Easy Level.}
    \label{fig:rand_easy}
\end{figure*}

\begin{figure*}[h]
    \centering
    \begin{subfigure}[b]{0.325\linewidth}
        \centering
        \includegraphics[width=\textwidth]{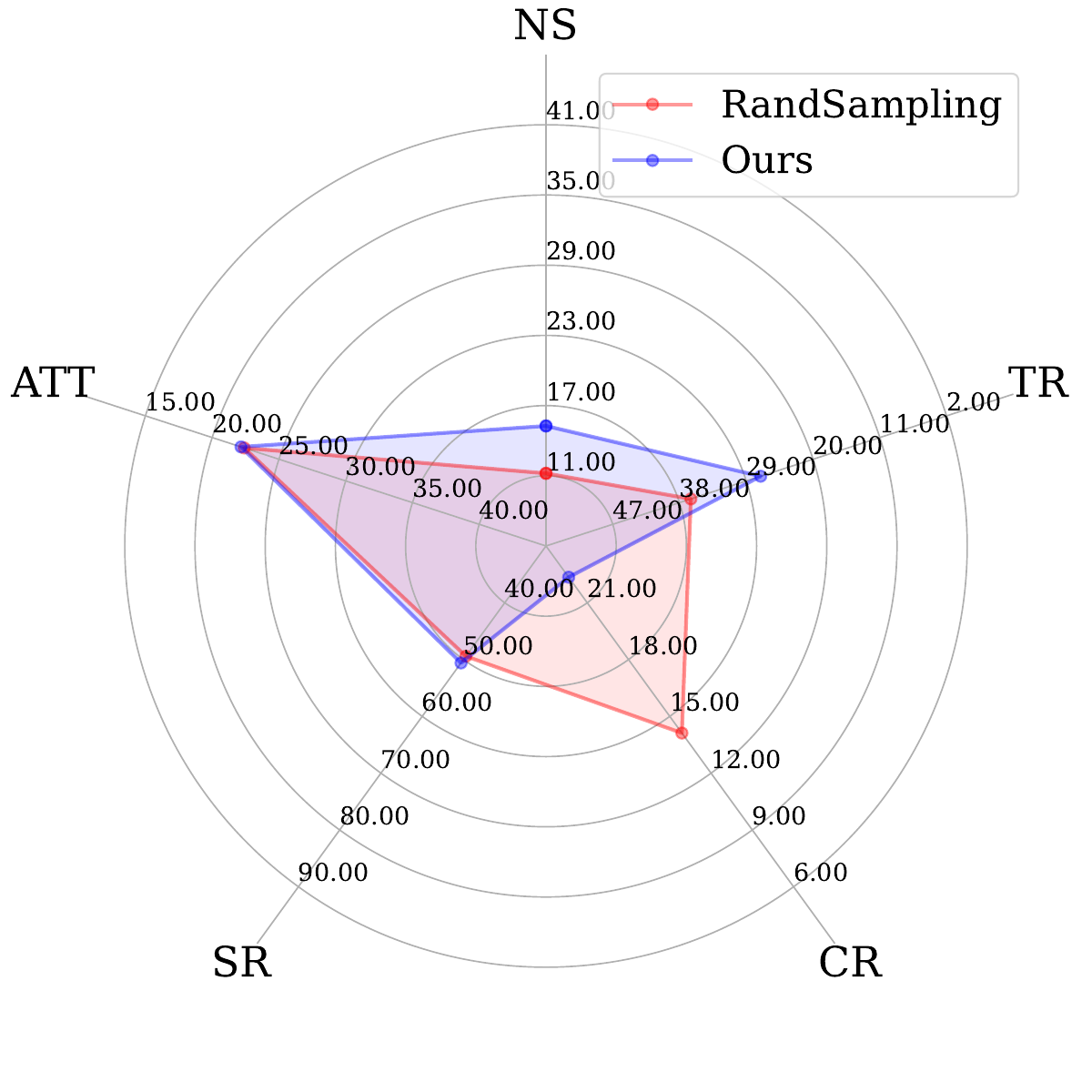}
        \caption{Train on Easy}
    \end{subfigure}
    \begin{subfigure}[b]{0.325\linewidth}
        \centering
        \includegraphics[width=\textwidth]{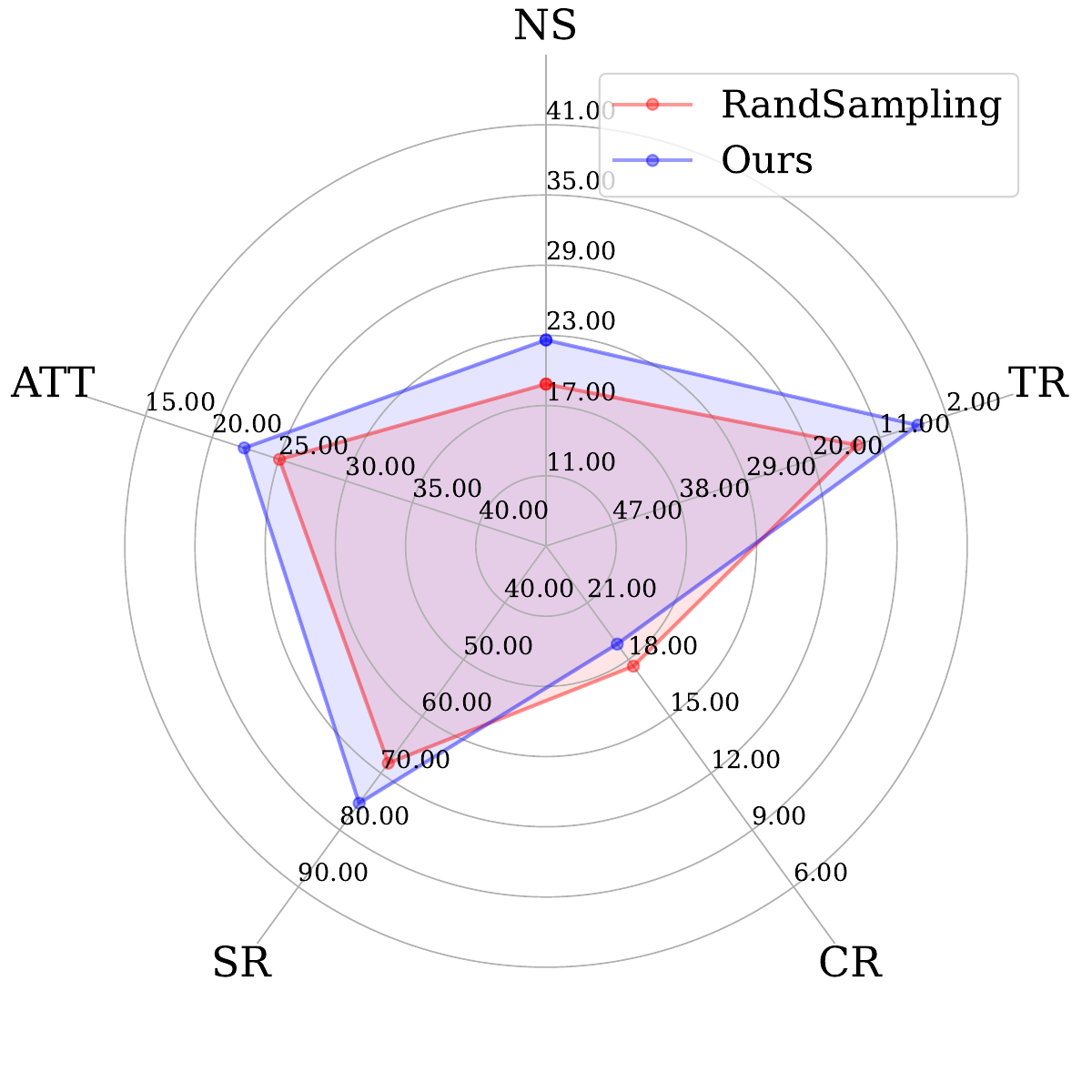}
        \caption{Train on Medium}%\label{fig:}
    \end{subfigure}
    \begin{subfigure}[b]{0.325\linewidth}
        \centering
        \includegraphics[width=\textwidth]{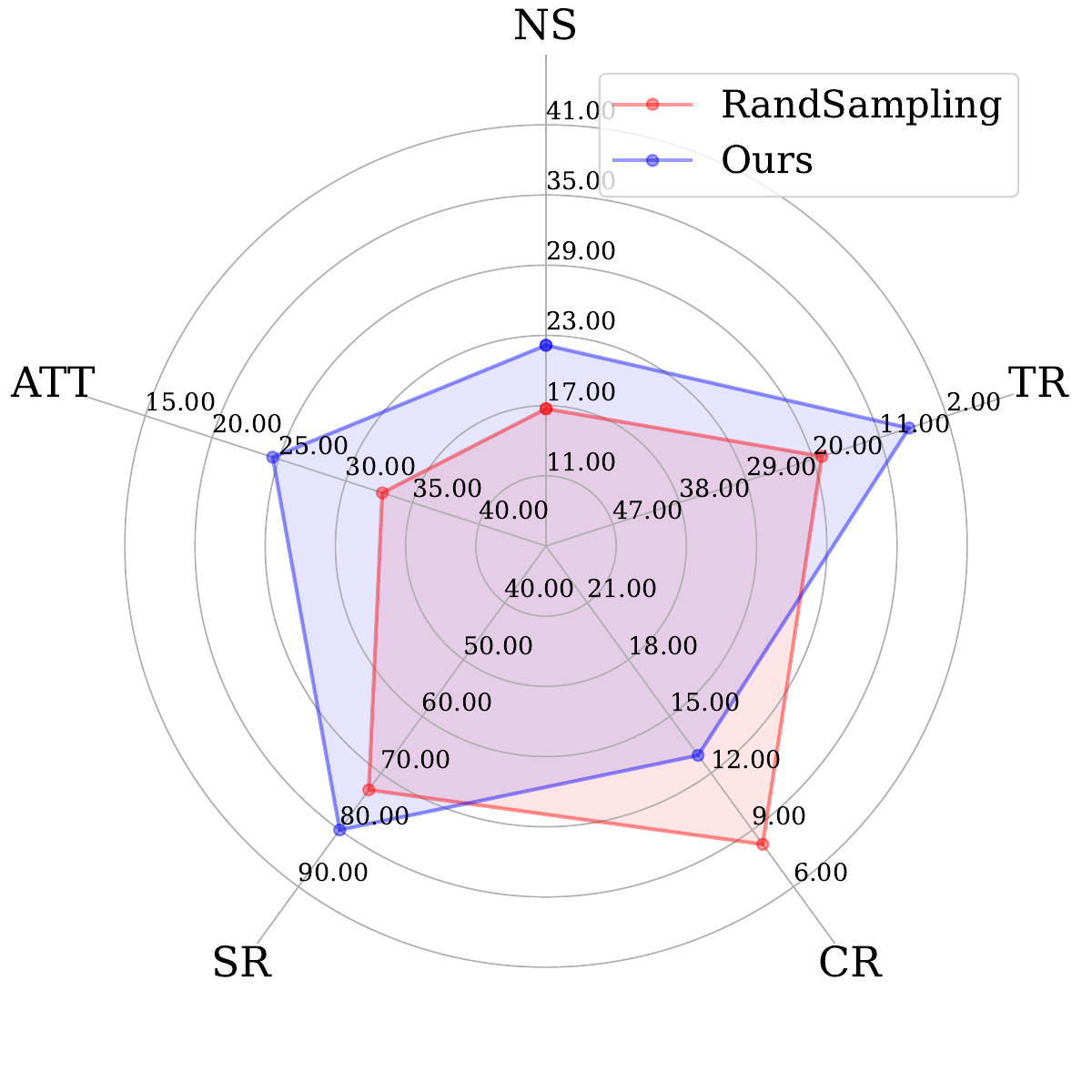}
        \caption{Train on Difficult}%\label{fig:}
    \end{subfigure}
    \caption{RandSampling v.s. \toolname. Test on Medium Level.\looseness=-1}
    \label{fig:rand_medium}
\end{figure*}

\begin{figure*}[!t]
    \centering
    \begin{subfigure}[b]{0.325\linewidth}
        \centering
        \includegraphics[width=\textwidth]{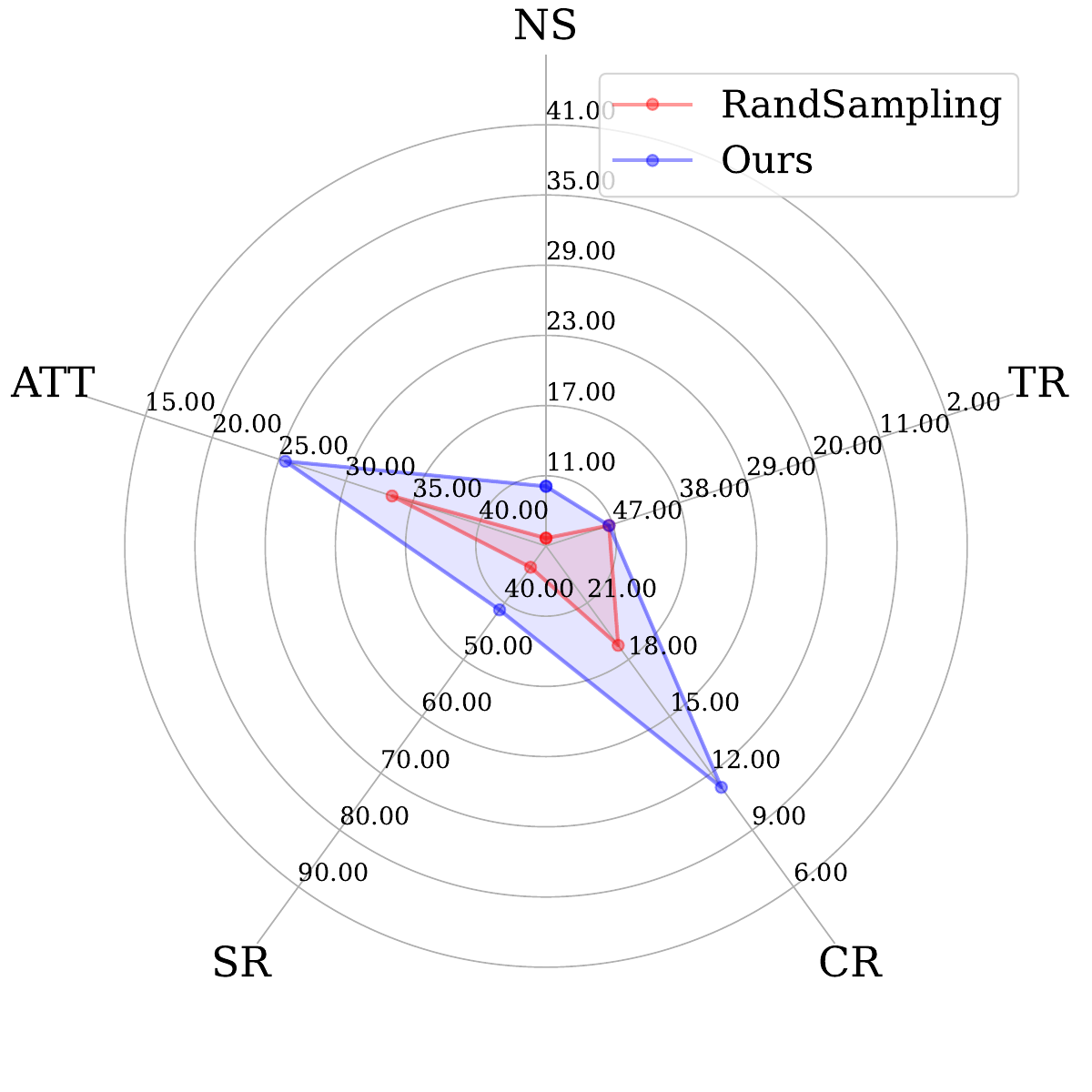}
        \caption{Train on Easy}%\label{fig:}
    \end{subfigure}
    \begin{subfigure}[b]{0.325\linewidth}
        \centering
        \includegraphics[width=\textwidth]{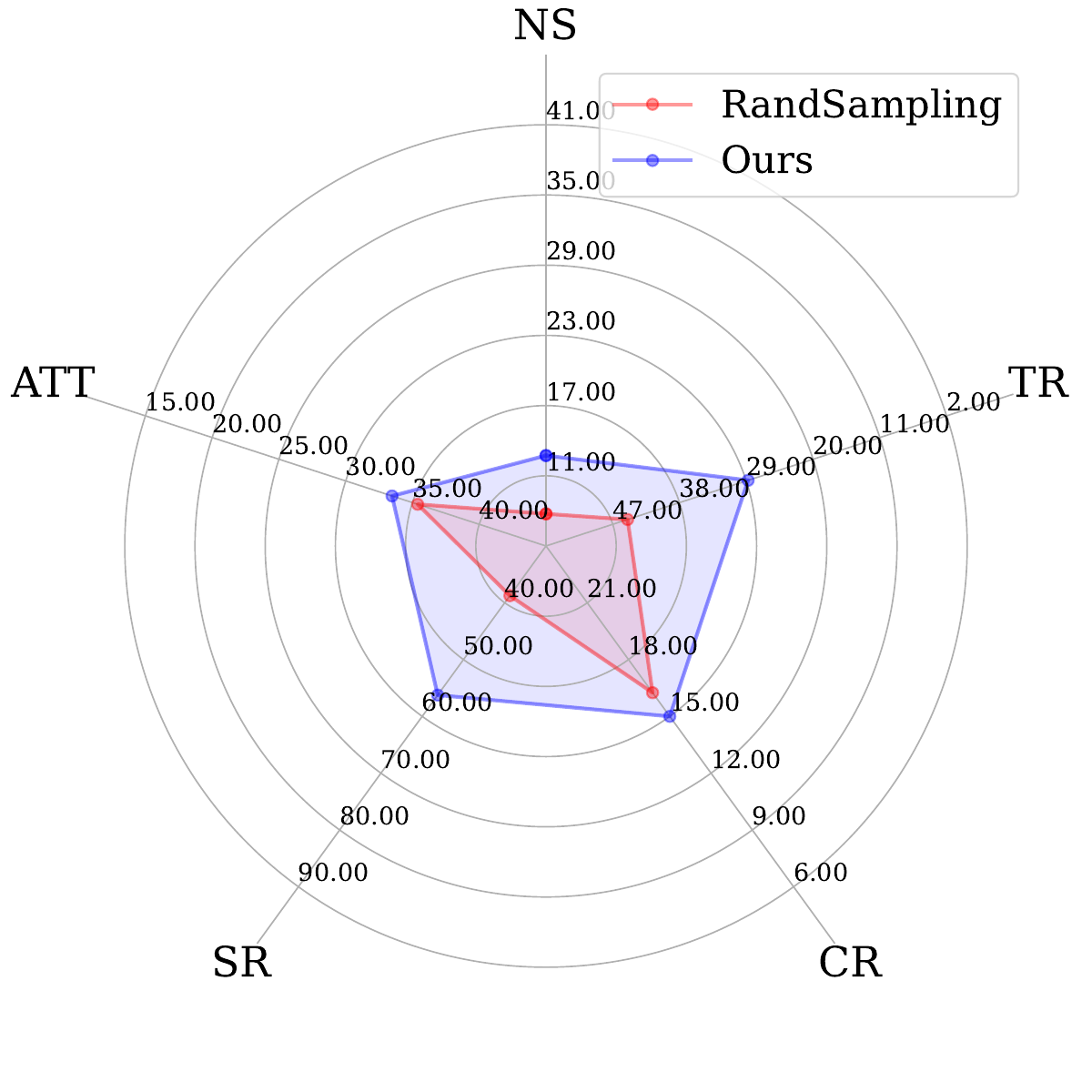}
        \caption{Train on Medium}%\label{fig:}
    \end{subfigure}
    \begin{subfigure}[b]{0.325\linewidth}
        \centering
        \includegraphics[width=\textwidth]{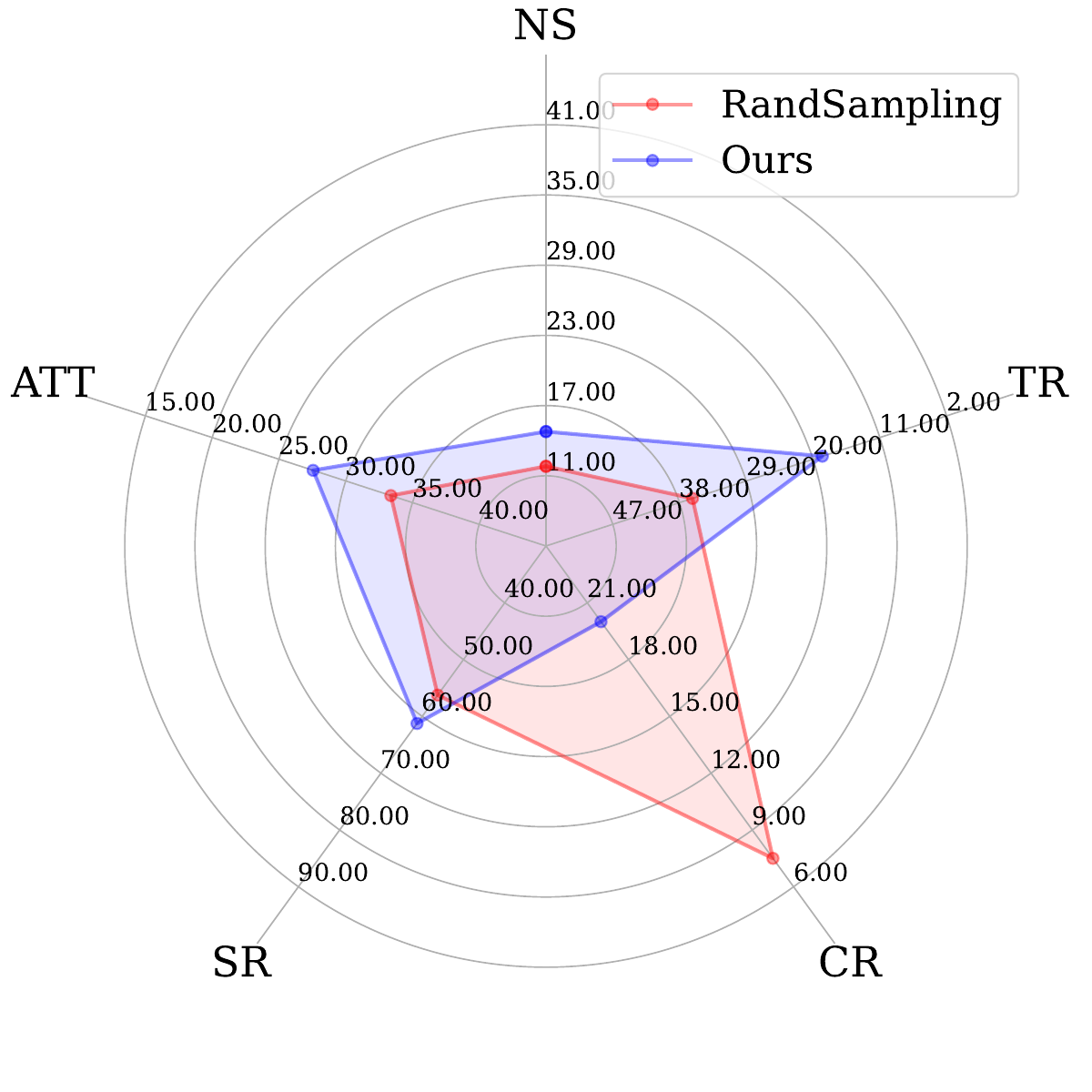}
        \caption{Train on Difficult}%\label{fig:}
    \end{subfigure}
    \caption{{\small RandSampling v.s. \toolname. Test on Difficult Level.}}
    \label{fig:rand_diff}
\end{figure*}

In this section, we delve deeper into the comparative analysis between RandomSampling and \toolname by applying both methods across environments of varying complexity: Easy, Medium, and Difficult.
We conducted separate experiments for each difficulty level, which allowed for detailed and direct comparisons. The results of these experiments are represented in Figures~\ref{fig:rand_easy},~\ref{fig:rand_medium}, and~\ref{fig:rand_diff}.

Consistent with the observations discussed in Section~\ref{sec:eval:diagnosis}, our \toolname continues to outperform RandomSampling in overall navigation performance. 
This superiority is quantified by a higher Navigation Score (NS), indicating \toolname's effectiveness in handling complex navigation tasks. 
This comparison not only highlights the strengths of \toolname but also contributes to a better understanding of its operational dynamics across different environmental settings. \looseness=-1

% \section{Additional Experiments} \label{sec:appx:additional_exp}

\begin{figure*}[t]
    \centering
    \begin{subfigure}[t]{.323\linewidth}
        \centering
        \includegraphics[width=1\linewidth]{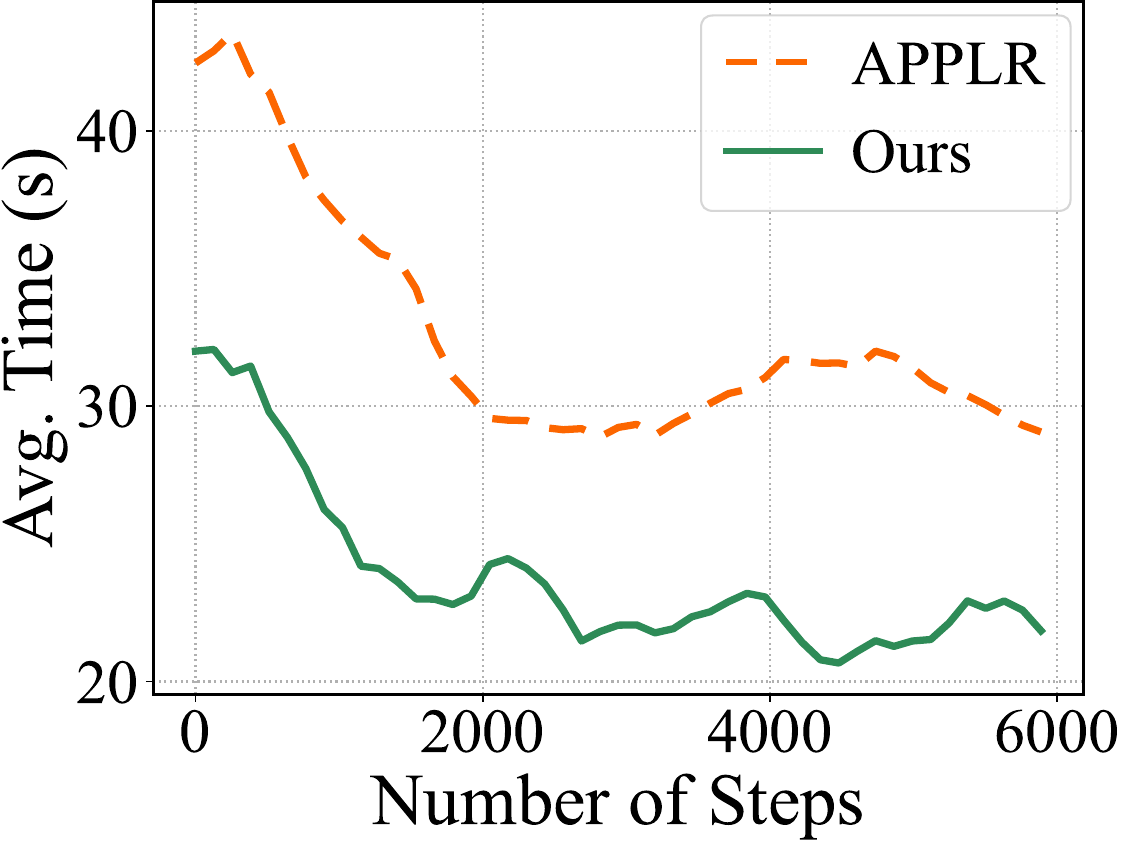}
        \caption{Average Time}
        \label{fig:eval:time_curve}
    \end{subfigure}
    \hfill
    \nextfloat
    \begin{subfigure}[t]{.325\linewidth}
        \centering
        \includegraphics[width=1\linewidth]{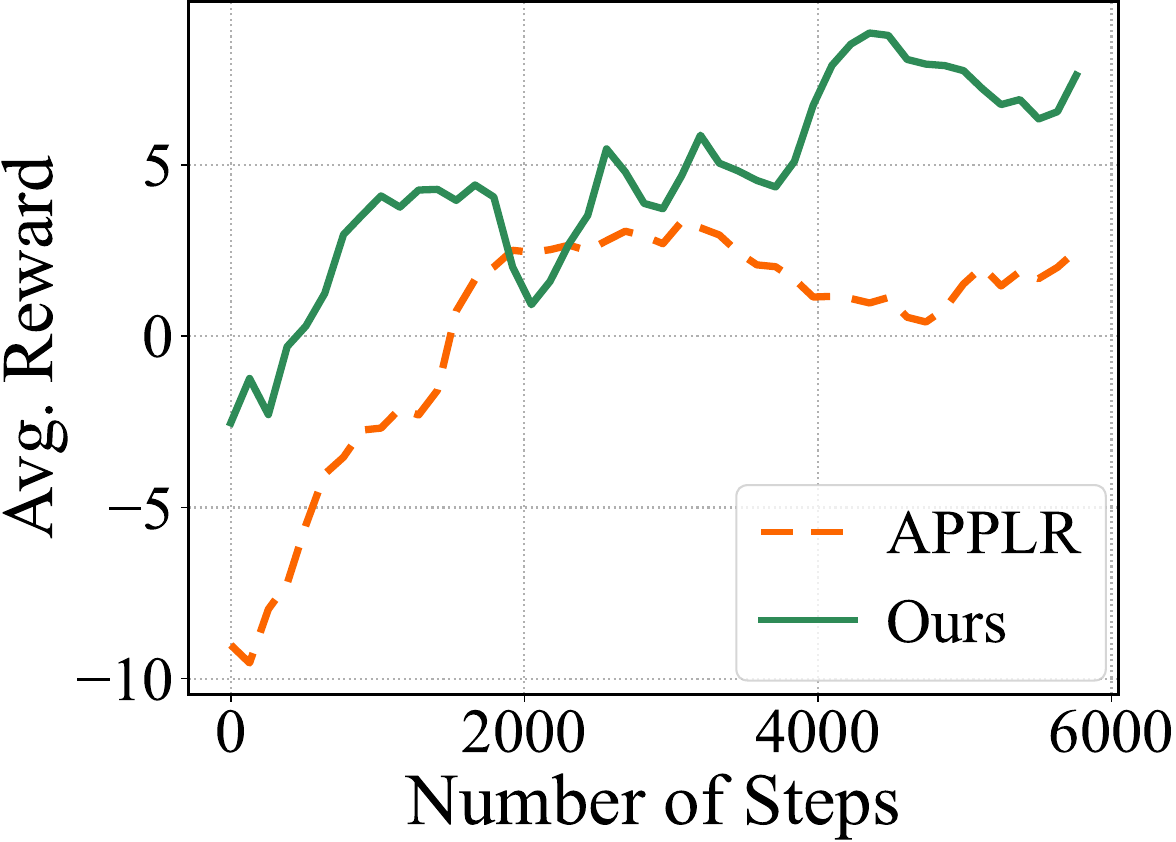}
        \caption{Average Reward}
        \label{fig:eval:reward_curve}
    \end{subfigure}
    \hfill
    \nextfloat
    \begin{subfigure}[t]{.325\linewidth}
        \centering
        \includegraphics[width=1\linewidth]{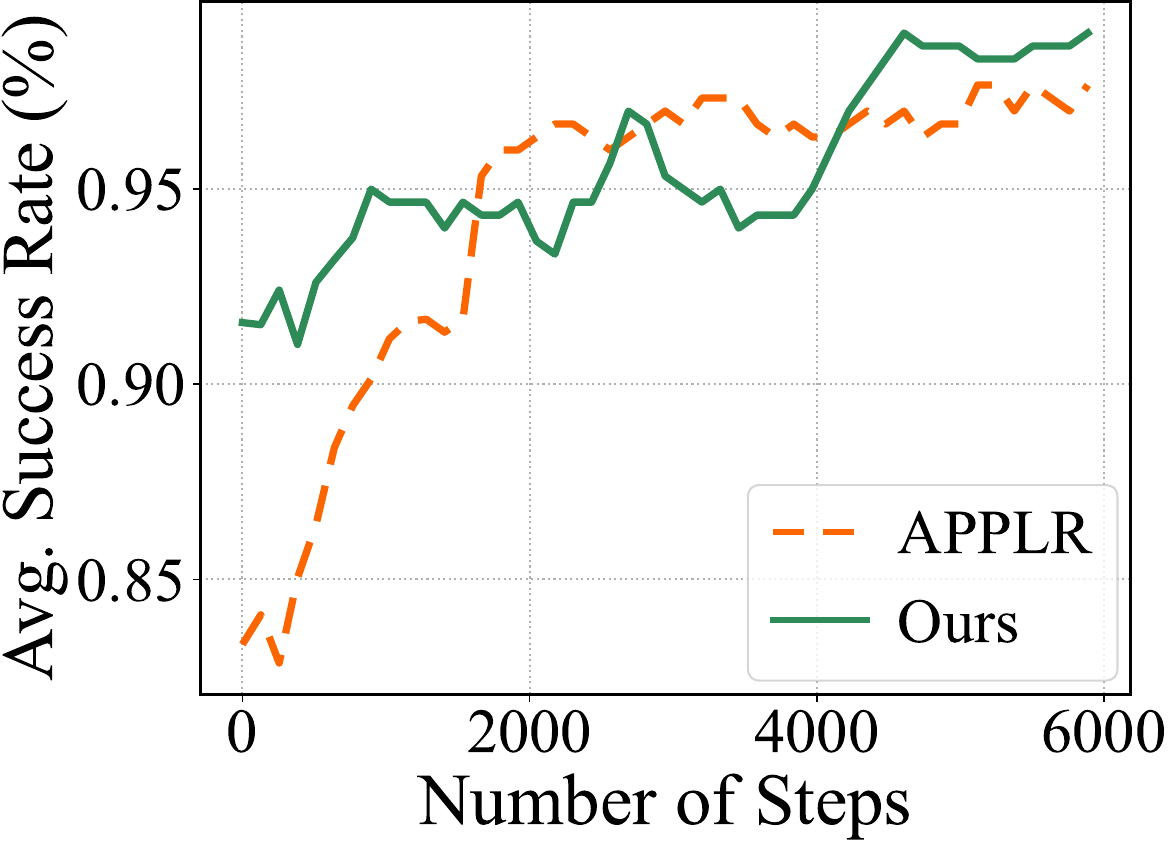}
        \caption{Average Success Rate}
        \label{fig:eval:success_curve}
    \end{subfigure}
    \caption{ Average time, reward, and success rate of \toolname and APPLR for different checkpoints during training.}
    \label{fig:eval:training_curve}
\end{figure*}

\begin{figure*}[h]
\begin{minipage}[b]{.49\textwidth}
\centering
%\captionsetup{belowskip=0pt,aboveskip=5pt}
\captionof{table}{Navigation performance under different high-resistance area up-sampling ratios $\lambda$.}
\resizebox{\textwidth}{!}{
\begin{tabular}{cccccc}
    \toprule
       Ratio $\lambda$ & NS $\uparrow$ & ATT $\downarrow$ & SR $\uparrow$ & CR $\downarrow$ & TR $\downarrow$ \\
    \midrule
      0.2 & 9.63 & \textbf{27.01} & 38.75 & 25.00 & 36.25 \\
      0.4 & \textbf{13.33} & 31.63 & \textbf{57.50} & \textbf{12.50} & \textbf{30.00} \\
      0.6 & 11.18 & 38.38 & 55.00 & 13.75 & 31.25 \\
      0.8 & 11.59 & 27.04 & 42.50 & 17.50 & 40.00 \\
    \bottomrule
\end{tabular}
\label{tab:eval:hard_sampling}
}
\end{minipage}
\hspace{5pt}
\begin{minipage}[b]{.48\textwidth}
\centering
%\captionsetup{belowskip=0pt,aboveskip=5pt}
\captionof{table}{ Navigation performance under different trajectory vector change thresholds $\eta$.}
\resizebox{\textwidth}{!}{
\begin{tabular}{cccccc}
    \toprule
       Threshold $\eta$ ($^\circ$)  & NS $\uparrow$ & ATT $\downarrow$ & SR $\uparrow$ & CR $\downarrow$ & TR $\downarrow$ \\
    \midrule
      50 & 15.19 & 26.89 & 50.59 & 18.82 & 30.59 \\
      70 & 18.92 & 26.78 & 65.20 & 14.80 & 20.00 \\
      90 & \textbf{22.61} & \textbf{22.40} & \textbf{75.29} & 18.82 & \textbf{5.88} \\
      110 & 18.22 & 28.93 & 74.12 & 15.29 & 10.59 \\
      130 & 17.45 & 23.70 & 57.65 & \textbf{14.12} & 28.24\\
    \bottomrule
    \end{tabular}
    \label{tab:eval:angle_threshold}
}
\end{minipage}
\end{figure*}

\section{More Results of Efficiency Comparison} \label{sec:appx:efficiency}
In Figure~\ref{fig:eval:training_curve}, we present additional metrics for various parameter tuner checkpoints during training, including average traveling time, reward, and success rate.
It clearly shows that \toolname outperforms APPLR across all three metrics.
Notably, \toolname requires fewer training steps to achieve shorter traveling times, highlighting the method's efficiency.
Similarly, within the same amount of training steps, \toolname can also achieve a higher average reward and success rate.
All of these three metrics robustly demonstrate that \toolname can achieve superior efficacy compared to the SOTA baseline APPLR.

\section{Ablation Study} \label{sec:appx:ablation}
\noindent \textbf{Sensitivity}. We conduct ablation studies to evaluate the
sensitivity of \toolname against various crucial hyper-parameters: (1) high-resistance area sampling threshold $\lambda$ in Algorithm~\ref{alg:main} and (2) high-resistance point identification threshold $\eta$ in Algorithm~\ref{alg:get_hr_area}. 

% \todo{check this ablation study result analysis: }
\noindent \textbf{Failure Trajectory Filtering}. We evaluate the necessity of our failure trajectory filtering in \toolname and its advantages for improving navigation performance. 
The results are shown in Table~\ref{tab:eval:failure_filtering}.
We observe that without the failure filtering trajectory mechanism, three out of five metrics deteriorate significantly, particularly the collision rate and success rate, which are critical to the robot's safety. 
In contrast, without failure filtering, the NS and ATT metrics show only marginal improvements. This indicates that filtering out failure trajectories can substantially enhance overall navigation performance in terms of safety and task completion rate while not sacrificing total travel time.

% \sw{The high-resistance area upsampling and original sampling should have a good balance.. } 
\noindent \textbf{High-resistance Area Up-sampling Ratio $\lambda$}. We then study the influence of $\lambda$ in Alg.~\ref{alg:main} on the navigation performance.
Table~\ref{tab:eval:hard_sampling} shows that $\lambda=0.4$ yields the best performance for the navigation system. 
Varying $\lambda$ introduces fluctuations across various metrics, while all within a 10\% range.
This demonstrates the necessity of fine-tuning the high-resistance area up-sampling ratio $\lambda$ to balance the proportion of difficult trajectories within the total training set.
If $\lambda$ is set too high, the training set will contain too many challenging trajectories, complicating the parameter tuner's ability to understand the complete map and complete the task. 
If $\lambda$ is set too low, it will revert to the original training method, hindering the parameter tuner's ability to learn from those under-fitted trajectories.

In Table~\ref{tab:eval:angle_threshold}, we show the performance metrics of \toolname over different threshold angles.
While there are some improvements in certain metrics (e.g., NS and CR) at higher threshold angles, the performance data clearly converges around 90° as the optimal threshold for overall system performance. 
This peak suggests at 90°, the system achieves a balance between speed, efficiency, and safety, highlighting a degree of robustness and adaptability. 
Similar to the high-resistance area up-sampling ratio $\lambda$, the threshold angle should also be carefully tuned to enable the best performance.
Thus, we set $\lambda=0.4$ and $\eta=90°$ for all other experiments.

\balance
\section{Potential Social Impacts}

The development and deployment of \toolname, aimed at enhancing the performance of parameter tuner policy in robotic navigation, promises significant positive social benefits. 
By enabling robots to navigate more effectively and reliably, our tool can greatly facilitate the integration of robotics into our daily life.
For example, in healthcare settings, more efficient navigation can enable robots to deliver medication, assist in surgeries, or provide companionship with greater precision and safety. 
In homes, robots equipped with advanced navigation capabilities can assist individuals with disabilities, offering them greater independence and quality of life. 
Furthermore, in disaster response situations, robots that can navigate challenging terrains could save lives by reaching areas that are inaccessible to humans. 
Overall, by improving the reliability and functionality of parameter tuners in robotic navigation systems, \toolname has the potential to make significant contributions to society, improving safety, accessibility, and efficiency in numerous critical areas.
%%%%%%%%%%%%%%%%%%%%%%%%%%%%%%%%%%%%%%%%%%%%%%%%%%%%%%%%%%%%%%%%%%%%%%%%%%%%%%%%

\end{document}